%% file: emnlp2021.tex
\pdfoutput=1
\documentclass[11pt]{article}

\usepackage{ACL2023}

\usepackage[algo2e]{algorithm2e}
\usepackage{times}
\usepackage{latexsym}
\usepackage{amsmath}

\usepackage{multirow}
\usepackage{makecell}
\usepackage[T1]{fontenc}
\usepackage[utf8]{inputenc}

\usepackage{microtype}
\usepackage{booktabs}
\usepackage{graphicx}
\usepackage{algorithm}
\usepackage{algorithmicx}
\usepackage{amsmath}
\usepackage{caption}
\usepackage{subfigure}
\usepackage{bbding}
\usepackage{color, soul, colortbl}

\newenvironment{itemize*}%
 {\leftmargini=12pt\begin{itemize}%
  \setlength{\itemsep}{5pt}%
  \setlength{\parskip}{0pt}%
  }%
 {\end{itemize}}
\newenvironment{enumerate*}%
 {\begin{enumerate}%
  \setlength{\itemsep}{0pt}%
  \setlength{\parskip}{0pt}}%
 {\end{enumerate}}

\definecolor{myred}{rgb}{0.839,0.341,0.271}
\definecolor{myblue}{rgb}{0.843,0.898,0.941}
\definecolor{mypurple}{rgb}{0.957,0.937,0.953}

\definecolor{myerror}{rgb}{0.996, 0.875,0.863}
\usepackage{tikz}
\usepackage{makecell}
\usepackage{cleveref}
\usepackage{ragged2e}
\crefname{section}{§}{§§}
\Crefname{section}{§}{§§}
\usepackage{tikz,lipsum}
\usepackage[most]{tcolorbox}
\usepackage{awesomebox} % for infobox
\usepackage[tikz]{bclogo}
\usepackage[framemethod=tikz]{mdframed}
\definecolor{bgblue}{RGB}{245,243,253}
\definecolor{ttblue}{RGB}{91,194,224}
\usepackage{enumitem}

\newtcolorbox{myboxi}[1][]{
  breakable,
  title=#1,
  colback=red!5,
  colbacktitle=red!5,
  coltitle=black,
  fonttitle=\bfseries,
  bottomrule=0pt,
  toprule=0pt,
  leftrule=2pt,
  rightrule=2pt,
  titlerule=0pt,
  arc=0pt,
  outer arc=0pt,
  colframe=red,
}

\title{Exploring Lottery Prompts for Pre-trained Language Models}

\author{Yulin Chen$^{1}$\thanks{\quad equal contributions}\hspace{0.5em}, Ning Ding$^{1,2*}$, {Xiaobin Wang}$^{3}$, \\  \textbf{Shengding Hu}$^{2}$, \textbf{Hai-Tao Zheng}$^{1,4\dag}$, \textbf{Zhiyuan Liu}$^{2,5,6}$\thanks{\quad corresponding authors}\hspace{0.5em}, \textbf{Pengjun Xie}$^{3}$ \\
$^{1}$Shenzhen International Graduate School, Tsinghua University $^{2}$DCST, Tsinghua University \\
$^{3}$Alibaba Group, 
$^{4}$Pengcheng Laboratory, Shenzhen, 
$^{5}$BNRIST, IAI, Tsinghua University  \\
$^{6}$Jiangsu Collaborative Innovation Center for Language Ability, \\Jiangsu Normal University, Xuzhou \\
\texttt{\{yl-chen21, dingn18\}@mails.tsinghua.edu.cn} \\
\texttt{\{zheng.haitao\}@sz.tsinghua.edu.cn}, 
\texttt{\{liuzy\}@tsinghua.edu.cn}}

\begin{document}
\maketitle
\begin{abstract}

\looseness=-1 Consistently scaling pre-trained language models (PLMs) imposes substantial burdens on model adaptation, necessitating more efficient alternatives to conventional fine-tuning.
% Prompting is found to be especially effective for few-shot and zero-shot settings. 
Given the advantage of prompting in the zero-shot setting and the observed performance fluctuation among different prompts, we explore the instance-level prompt and their generalizability.
% While many works have focused on prompt engineering and optimization, few work has paid attention to instance-level prompt mining for PLMs.
By searching through the prompt space, we first validate the assumption that for every instance, there is almost \textit{always} a lottery prompt that induces the correct prediction from the PLM, and such prompt can be obtained at a low cost thanks to the inherent ability of PLMs.
Meanwhile, we find that some strong lottery prompts have high performance over the whole training set, and they are equipped with distinguishable linguistic features.
% We also show that the cost needed to search for a lottery prompt decreases with the pre-training of PLMs.
Lastly, we attempt to generalize the searched strong lottery prompts to unseen data with prompt ensembling method without any parameter tuning.
Experiments are conducted on various types of NLP classification tasks and demonstrate that the proposed method can achieve comparable results with other gradient-free and optimization-free baselines.
%under few-shot setting. 
%We also analyze the effect of ensembling strategy and training data size on the test set performance.
\end{abstract}

\input{intro.tex}

\input{existence.tex}

\input{generalization.tex}

% \vspace{-0.1cm}
\section{Related Work}
\looseness=-1 Prompting, as an alternative to standard finetuning, is originally inspired by GPT-3~\citep{brown2020language} and knowledge probing~\citep{petroni-etal-2019-language, jiang-etal-2020-know}. With a similar form to pre-training tasks, it stimulates the intrinsic knowledge in PLMs more efficiently. 
Following several of the earliest works~\citep{schick2021exploiting, schick2021just}, prompting has been applied in various NLP tasks~\citep{han2021ptr, li2021prefix, sainz2021label, ding2021prompt}.
It is also discovered that the specific prompt used has a great impact on task performance. Therefore, efforts have been devoted to prompt engineering and automatic prompt generation. Optimizing for a good prompt has been conducted at both discrete token level~\citep{shin-etal-2020-autoprompt, gao-etal-2021-making} and continuous embedding level~\citep{li2021prefix, zhang2021differentiable, liu2021gpt, li-etal-2022-spe}. Some also focus on the choice and representation of label words~\citep{schick-etal-2020-automatically, hu2021knowledgeable, zhang2021differentiable}. Experiments show that a well-optimized or pre-trained~\citep{gu-etal-2022-ppt} prompt can be beneficial.

Given the striking performance of prompting under few-shot settings especially, recently, more works are focusing on the more efficient tuning of PLMs based on prompts. Prompt tuning~\citep{lester2021power} tunes the pre-pended token embedding only. %P-Tuning v2~\citep{liu2021ptuning} extends the concept of prompt tuning to every layer of the transformer structure. 
Other works enhance PLMs' zero-shot learning ability with prompts. 
Studies show that large PLMs with proper prompts~\citep{wei2021fintuned} and training with diverse prompts~\citep{sanh2021multi} can advance zero-shot performance. This line of work emphasizes the efficient tuning and steering process of large PLMs.
%Aside from tuning only a few parameters, a further approach is optimization in a gradient-free manner. 
Black-box tuning~\citep{sun2022black} optimizes the pre-pended continuous prompt in a projected low-dimensional space without PLM gradient information.

\looseness=-1 This work is among the first few efforts~\citep{jin2022instance, wu2022idpg} in mining instance-level prompts, and is the first to propose and prove the existence of a lottery prompt composed of a few textual tokens for each instance. In contrast to tuning a small number of parameters or tuning without gradients, an optimization-free method is proposed to generalize the searched prompts to the test sets.

% \vspace{-0.05cm}
\section{Conclusion}
% \looseness=-1 
In this work, we explore the existence of lottery prompts for every single instance and the adaptation of them for various classification tasks in an optimization-free manner. We propose a large prompt space composed of common words as the search space to verify the assumption. We also identify the searched strong prompts and the relation between model capacity and search cost and demonstrate the effectiveness of ensembling the strong prompts on the test set. Our proposed optimization-free method achieves satisfactory results on various NLP tasks under few-shot settings. 
Above all, this work illuminates the fact that the great potential of PLMs can be successfully harnessed and prompted by plain textual prompts mined from PLM vocabulary without parameter optimization and thus points to the need for future efforts in more efficient ways in mining and utilizing lottery prompts. 

\section*{Acknowledgements}
This research is supported by 
the National Natural Science Foundation of China (Grant No.62276154 and No.62236004), Research  Center for Computer Network (Shenzhen) Ministry of Education, Beijing Academy of Artificial Intelligence (BAAI), the Natural Science Foundation of Guangdong Province (Grant No. 2023A1515012914), Basic Research Fund of Shenzhen City (Grant No. JCYJ20210324120012033 and JSGG20210802154402007), the Major Key Project of PCL for Experiments and Applications (PCL2021A06), Overseas Cooperation Research Fund of Tsinghua Shenzhen International Graduate School (HW2021008), Major Project of the National Social Science Foundation of China  (No. 22\&ZD298), and Institute Guo Qiang at Tsinghua University. Finally, we thank Xiaozhi for providing valuable comments.

\section*{Limitations}
The current method works with a large prompt search space $\mathcal{T}$, which means a tremendous number of inference API calls are required. Though Figure~\ref{fig:search} shows that the average cost of finding a lottery prompt for each instance is low, the searching process is highly randomized and there is no guarantee of the performance of searched prompts over the test dataset. Therefore, finding strong prompts over the training set can still be laborious. 
How to use PLM inference calls more efficiently and leverage the generalization ability of $\mathcal{T}^*$ within a reasonable cost is of future research interest.
Our acceleration strategy can be found in Appendix~\ref{sec:efficiency}.

Another aspect is that not all strong prompts are interpretable as presented in \ref{tab:top_prompt_example}. While recently emerged larger models like ChatGPT demonstrate superb language understanding ability and can almost always answer yes or no questions correctly given a human-interpretable prompt. This gap observed between small PLMs like RoBERTa and large language models like ChatGPT is yet another interesting research topic.

\section*{Ethical Considerations}
This work shows that with proper plain textual prompts, instance-level desired results can be prompted from PLMs. 
This inherent feature of PLMs means attacks can be launched to produce rude or discriminated words.
On the other hand, however, we believe it can also be a technique used for debiasing a PLM. Overall, this effect depends on the intention of the users and the pre-training corpus of the corresponding PLMs. The analysis of this study can be used to facilitate the community to develop more specifications for the rational use of PLMs (especially the super-large ones), and more approaches to effectively prevent potential ethical issues. For example, we can use this technique to analyze which outputs that may have ethical issues are easily triggered by which contexts (prompts) and develop a set of intervention methods to make these tokens unavailable for output.

\bibliography{anthology,custom}
\bibliographystyle{acl_natbib}

\clearpage
\appendix
\input{appendix.tex}

\end{document}

%% file: intro.tex
\section{Introduction}

\looseness=-1 Since pre-trained language models (PLMs) became the de-facto standard in modern NLP researches~\cite{devlin2018bert, liu2019roberta, HAN2021}, the pretraining-finetuning paradigm has been prevailing until recent years when models keep scaling~\cite{radford2019language, brown2020language, rae2021scaling} and become too expensive to be optimized. 
To this end, researchers are actively seeking more effective strategies that require little or even no optimization to harness PLMs. 

%In more extreme cases, users can only access the inference results of extremely large PLMs and are not able to make optimization-based adaptations to downstream tasks. To this end, alternative methods for model adaptation that requires little or even no optimization are urgently needed and actively explored. 

 %Researchers began seeking for more efficient and cost-effective approaches to adapt PLMs to their downstream tasks.
\looseness=-1 Among these exploratory studies of advanced model adaptation, prompting~\cite{brown2020language, schick-etal-2020-automatically, schick2021exploiting, gao-etal-2021-making} is gaining popularity in the community, which uses additional context (prompts) to wrap input instances and trigger desired output tokens. Note that in this paper, the term ``prompt'' technically refers to the template that wraps the original input. In classification tasks, these tokens are further mapped to particular labels by a verbalizer. Such a paradigm is verified to be effective in a variety of downstream tasks, even when annotations are insufficient. 
Particularly, empirical evidence shows that coincidental prompts could achieve extraordinary performance in the zero-shot setting, i.e., no training examples are presented. % for example 可以加到这里
For example, simple manual prompt can achieve an F1 score of over 60\% on 46-class entity typing dataset~\citep{ding2021prompt} and reaches 73\% accuracy on DBpedia with 14 classes~\citep{hu2021knowledgeable} in the zero-shot setting.

\looseness=-1 Despite the promising performance of prompting, it is often accompanied by drastic fluctuations among different prompts~\cite{zhao2021calibrate}.
% Although such effectiveness is often accompanied by drastic fluctuations with prompt selection~\cite{zhao2021calibrate}, it still implies that prompting PLMs may implicitly have an exceedingly high upper capability even without model tuning. 
% And it may be possible to tap into the potential by assigning different prompts for distinct data points. % need polish
Given the observed sensitivity and context-dependent nature of the prompting method, it is intuitive to assign distinct prompts to each instance to trigger the desired output. % need polish
Intrigued by this intuition, we explore a bold hypothesis: 

% \begin{myboxi}[Hypothesis]
% Is it possible to find at least one instance-level prompt that induces correct output for every data point (lottery prompt) in classification tasks without any optimization?
% \end{myboxi}
\begin{tcolorbox}[
% title=#1,
%   colback=white,
  colback=gray!5,
  colbacktitle=gray!5,
  coltitle=black,
  fonttitle=\bfseries,
  bottomrule=0.5pt,
  toprule=0.5pt,
  leftrule=0.5pt,
  rightrule=0.5pt,
  titlerule=0pt,
  arc=0pt,
  outer arc=0pt,
  colframe=gray]
\textit{Is it possible to find at least one instance-level prompt that induces correct output for every data point (lottery prompt) in classification tasks without any optimization?}
\end{tcolorbox}

We empirically show that after building an automatic searching procedure with reasonable searching space on 13 representative classification datasets of up to 66 classes, \textbf{the existence of such lottery prompts can be validated (\cref{sec:existence}) }. That is, the combination of just a few discrete tokens can make a PLM output correct labels for almost any classification data. This finding updates our recognition of the limit of prompted knowledge in PLMs and demonstrates a promising upper bound of the PLMs' inference capability.

\looseness=-1 With the hypothesis verified, we conduct further analysis on the internal mechanisms and properties of lottery prompts to explore how the lottery prompts relate to model capability and how lottery prompts generalize to unseen data without any optimization. 

(1) We first find that the search cost of lottery prompts is low for most datasets (under 30 API calls), and could reflect task difficulty and model capacity~(\cref{sec:search_cost}). Search success rate increases and search cost decreases for larger PLMs and PLMs pre-trained for more steps, demonstrating that lottery prompts are a unique consequence of the expanded model capacity, rather than a mere stroke of luck. 
(2) Among these lottery prompts, we also find that there are a number of ``strong prompts'' that perform non-trivially on the whole training set, and interpretable linguistic features can be identified among them~(\cref{sec:strong_prompt}). 
% That is, although one ``strong prompt'' is obtained by searching for just one data point, it can be used to facilitate prediction on the entire dataset. 
Strong prompts demonstrate considerable potential to be generalized to unseen data, i.e., test dataset, of the current task. We develop a mutual-information-based prompt ensembling method and show that strong prompts could be effectively generalized to unseen data in an optimization-free manner~(\cref{sec:generaliza}). Without any parameter update, the ensembling of strong prompts could achieve on-par or better performance with many competitive baselines. 

In summary, we validate the existence of lottery prompts and conduct an in-depth analysis of the properties of lottery prompts. 
% Our findings reveal the relations between lottery prompts, data difficulty and PLM capacity, and improve our understanding of the upper limit of PLMs with prompting.
We also show that by directly ensembling the strong prompts, prominent performance can be achieved on test data without any optimization. Our study points to the great potential of PLMs and is hoped to inspire future works in more efficient ways in searching and ensembling lottery prompts as an optimization-free adaptation of PLMs.

%% file: existence.tex
\section{The Existence of Lottery Prompts for Every Data Point}
\label{sec:existence}
% \vspace{-0.2cm}

Considering the extraordinary performance observed on zero-shot classification and the large variance brought by the prompt selection, we make an assumption as follows: {Given a pre-trained language model and a classification dataset, for each instance, at least one lottery prompt exists that can induce the desired label from the PLM, without the need to update the PLM parameters.} 

To validate the assumption, we conduct experiments that attempt to find the lottery prompt for every data point on 13 classification tasks. Note that for different instances, the prompt may be different, and our goal is to verify the existence of such prompts in this experiment.

\subsection{Overview and Setup}

\looseness = -1 Particularly, for every input instance in a classification task, we attempt to search through the prompt space and find a textual prompt that can make PLMs produce desired label words.
We choose 13 datasets of various NLP tasks for assumption validation. Most of them come from GLUE benchmark~\citep{wang2018glue}, and others include Yelp Polarity~\citep{zhang2015character}, SNLI~\citep{bowman-etal-2015-large}, AG's News~\citep{zhang2015character}, DBpedia~\citep{zhang2015character}, and Few-NERD~\citep{ding-etal-2021-nerd}. 
SST-2~\citep{socher-etal-2013-recursive} and Yelp Polarity are datasets for binary sentiment classification. 
CoLA~\citep{warstadt2018neural} is for acceptibility judgment of single sentence.
SNLI, RTE~\citep{wang2018glue}, QNLI~\citep{wang2018glue}, WNLI~\citep{levesque2011logical} and MNLI~\citep{williams-etal-2018-broad} target at language inference detection given a sentence pair. 
QQP~\citep{iyer2017first} and MRPC~\citep{schick-etal-2020-automatically} are for paraphrase judgment.
AG's News and DBpedia are used for text theme classification. 
Few-NERD is an entity typing dataset. 

As for prompt search space, 200 words with top frequency in English\footnote{\url{https://sketchengine.co.uk}} are gathered and grouped according to part-of-speech tag with NLTK package~\citep{loper2002nltk} into nouns, verbs, prepositions, adjectives and adverbs. The designed prompt search space is the Cartesian product of three word sets $\mathcal{T}=\texttt{NOUNS} \times \texttt{VERBS} \times (\texttt{PREP}\cup \texttt{ADJ} \cup \texttt{ADV}) \times \{\texttt{<MASK>}\}$, and $|\mathcal{T}|=76725$. The major concerns of such designing is to restrict the prompt space and to fit with common syntactic order of words to ensure prompt plausibility to some extent. As for verbalizers, we follow the standard design of previous works~\citep{sun2022black}.
We use RoBERTa-large~\citep{liu2019roberta} and GPT-2~\citep{radford2019language} as the backbones.
The specific prompt format and verbalizers used are shown in Appendix~\ref{sec:promptappendix}.

\subsection{The Searching Process}
\looseness=-1 For each dataset, we randomly sample 1000 instances from the training set as $\mathcal{X}_\text{train}=\{(x_i, y_i)\}$ and apply each prompt $T \in \mathcal{T}$ to each instance and use the PLM $\mathcal{M}$ to produce the prediction.
Specifically, a prompt $T$ composed of a noun, a verb and an adjective may be \textit{``it was really''}. Applying it to an instance $x$:\textit{``A fun movie.''} will yeild the input text $T(x)$:\textit{``A fun movie. it was really \texttt{<MASK>}''}.
%%%%%%%%%%%%%%%%%%%%%%%%%%%% explanation needed
% as shown in Figure~\ref{fig:example}. 
For each of such pair $T(x) \in \mathcal{X}_\text{train} \times \mathcal{T}$, the score for each class can be obtained as 
\begin{equation}
    o(x;T,\mathcal{M}) = \texttt{Softmax}(\text{V}(\mathcal{M}(T(x)))),
\end{equation}
\looseness=-1 where \text{V} denotes the projection from output logits over PLM vocabulary to the class label set.
Specifically, to reduce the impact from the prompt, we use calibration~\citep{zhao2021calibrate} to rescale the scores before making the final prediction. 
\begin{equation}
\begin{aligned}
    q(T;\mathcal{M}) &= \texttt{Softmax}(\text{V}(\mathcal{M}(T(\cdot)))), \\
    p(x;T, \mathcal{M}) &= \texttt{Normalize}(\frac{o(x;T,\mathcal{M})}{q(T;\mathcal{M})}).
\end{aligned}
\label{eq:p}
\end{equation}

$T(\cdot)$ means a wrapped input with empty string and $q$ is the corresponding output probability over the label words. $p$ is the final calibrated probability over the class labels. 
For every $(x, y) \in \mathcal{X}_\text{train}$, we enumerate over each $T \in \mathcal{T}$ and see if the output $\hat{y}=\arg\max{p}$ will give the correct prediction $y$.

\subsection{Verification of the Assumption} Table~\ref{tab:stage1} reports the basic searching results. Each instance $x$ is considered correctly predicted if there exists $T \in \mathcal{T}$ such that $y= \arg\max {p}$. It is shown that for all datasets, a lottery prompt that induces the correct prediction from $\mathcal{M}$ exists for almost all 1000 instances. 
The assumption is thus validated, that is, in a finite search space composed of textual tokens, 
we can almost always find at least one combination of common words as a prompt to make the prediction correct. 
While it may not be surprising to see a success on binary classification tasks, achieving 100\% coverage on Few-NERD, a 66-class dataset for entity typing, is worth noting.
It indicates that the particular semantics distributed in PLM can be triggered by certain contexts even without any further fine-tuning.
% %%%%%%%%%% further illustrate 

Naturally, the phenomenon is not observed when the model is not pre-trained. We conduct the same searching process for Few-NERD on a randomly initialized RoBERTa-large, and only 33.1\% instances could find the corresponding lottery prompts. The effect of pre-training will be further explored in Section \ref{sec:pretrain}, demonstrating that lottery prompts are a unique and consequent effect along with language model pre-training.

\begin{table}[!ht]
    \centering
    \scalebox{0.86}{
    \begin{tabular}{lccc}
        \toprule
        \textbf{Datasets} &\#Classes & {RoBERTa-large} & GPT-2  \\ \midrule
        \textbf{SST-2} &2& 100.00 & 100.00  \\
        \textbf{Yelp P.} & 2 & 100.00 & 100.00  \\
        \textbf{SNLI} & 3 & 100.00 & 99.90  \\
        \textbf{RTE} & 2 & 100.00 & 100.00  \\
        \textbf{MRPC} & 2 & 100.00 & 100.00  \\
        \textbf{CoLA} & 2 & 100.00 & 100.00  \\
        \textbf{MNLI} & 3 & 99.90 & 99.90 \\
        \textbf{QNLI} & 2 & 100.00 & 100.00  \\
        \textbf{QQP} & 2 & 100.00 & 100.00  \\
        \textbf{WNLI} & 2 & 100.00 & 100.00  \\
        \textbf{AG's News} & 4 & 100.00 & 100.00  \\
        \textbf{DBpedia} & 14 & 100.00 & 100.00 \\
        \textbf{Few-NERD} &66 & 100.00 & 99.70  \\
        \bottomrule
    \end{tabular}}
    \caption{\looseness=-1 The success rate (\%) of lottery prompt search for each dataset's 1000 randomly sampled data. WNLI uses the whole training set with 635 instances.}
    \label{tab:stage1}
    % \vspace{-0.5cm}
\end{table}
% \vspace{-0.5cm}

\section{Empirical Analysis}
\label{sec:emp_ana}
Since we have verified the existence lottery prompts, in this section, we conduct further analysis on search cost and the searched lottery prompts.
% to explore the internal interaction between lottery prompts, data and PLMs.
% to answer the first three research questions.

\begin{figure*}[!htbp]
\captionsetup[subfigure]{labelformat=empty} 
\centering
\subfigure{
\begin{minipage}[t]{0.47\linewidth} 
\centering
\includegraphics[width=1.0\linewidth]{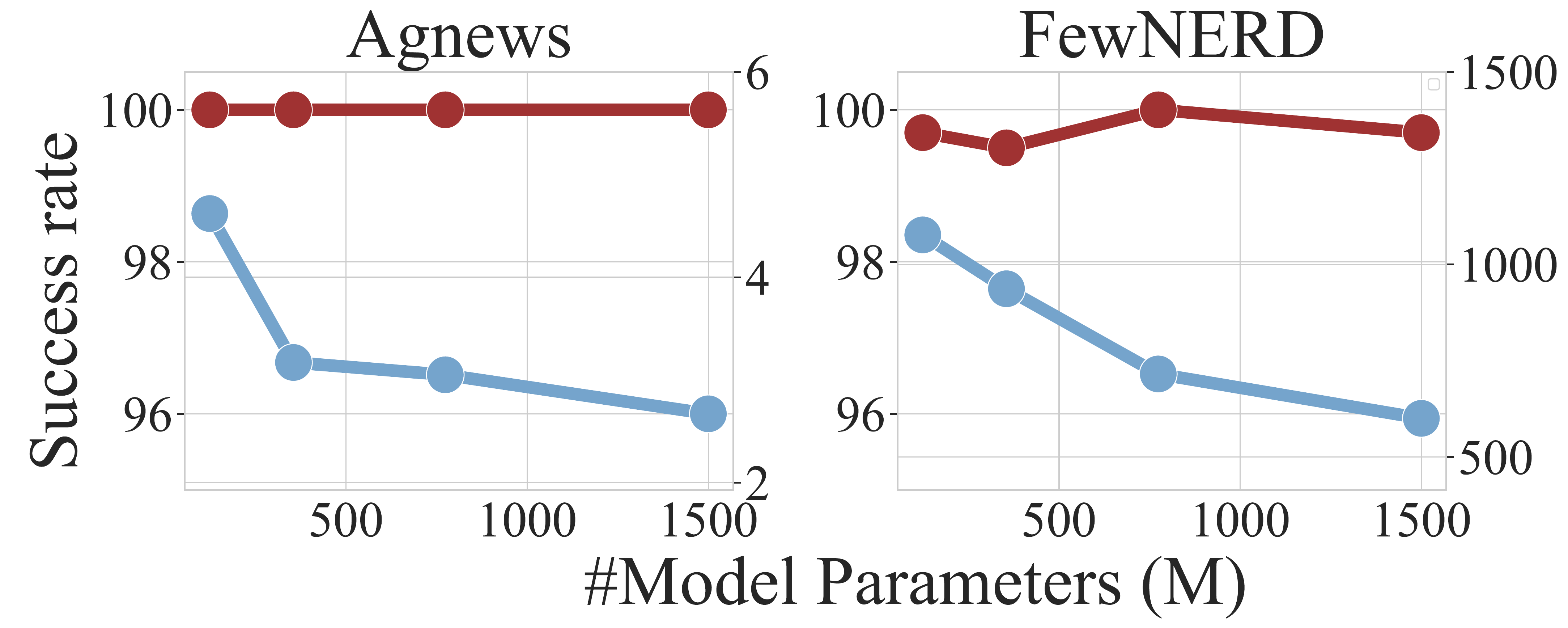} 
%\caption{fig1}
\end{minipage}%
}%
\hspace{0.3cm}
\subfigure{
\begin{minipage}[t]{0.49\linewidth} 
\centering
\includegraphics[width=1.0\linewidth]{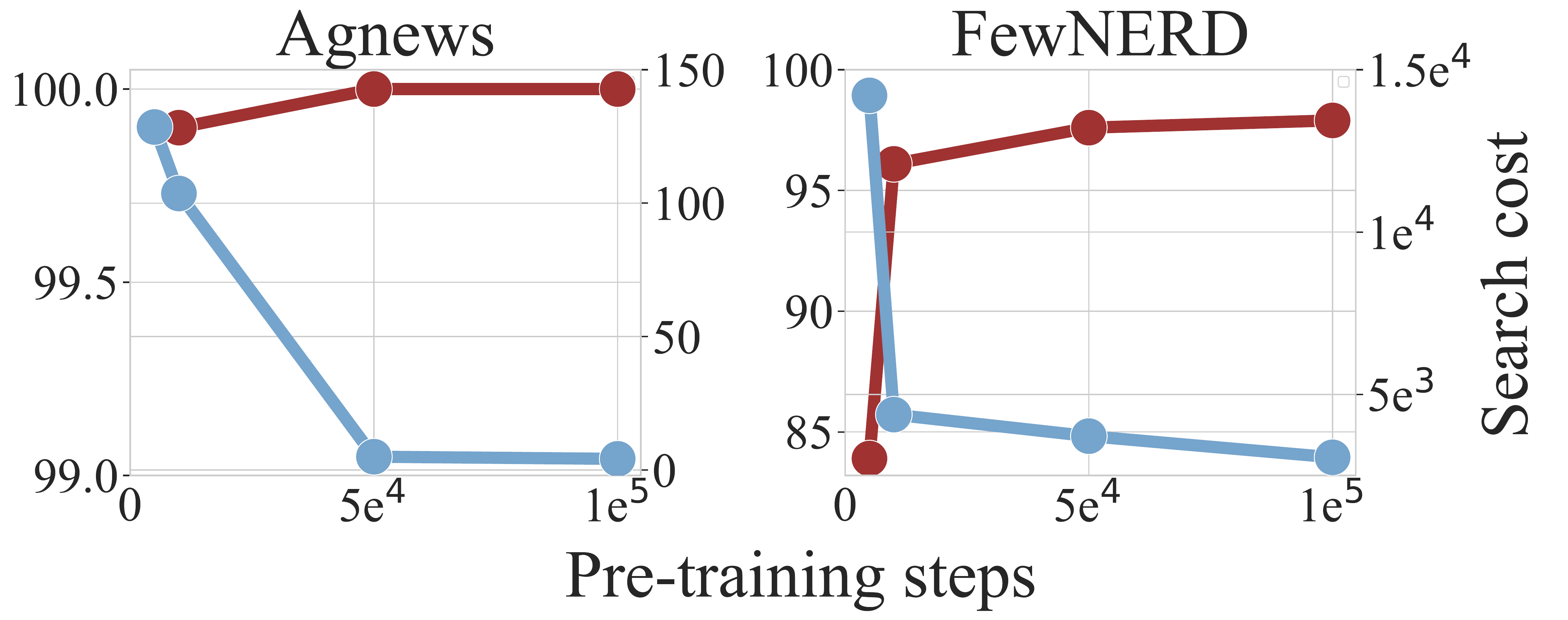}
%\caption{fig2}
\end{minipage}%
}%

\centering
% \vspace{-0.2cm}
\caption{The change of search success rate and the mean search cost along with the model size (left) and the pre-training steps (right). Experiments are conducted with GPT-2, GPT-2-medium, GPT-2-large, GPT-2-xl, and RoBERTa-base pre-trained with 5000, 10000, 50000, and 100000 steps.}
\vspace{-0.2cm}
\label{fig:pretrain}
\end{figure*}

\begin{figure}[!htbp]
\centering
    % \subfigure[]{
	\begin{minipage}{0.98\linewidth}
		\centering
		\includegraphics[width=0.9\linewidth]{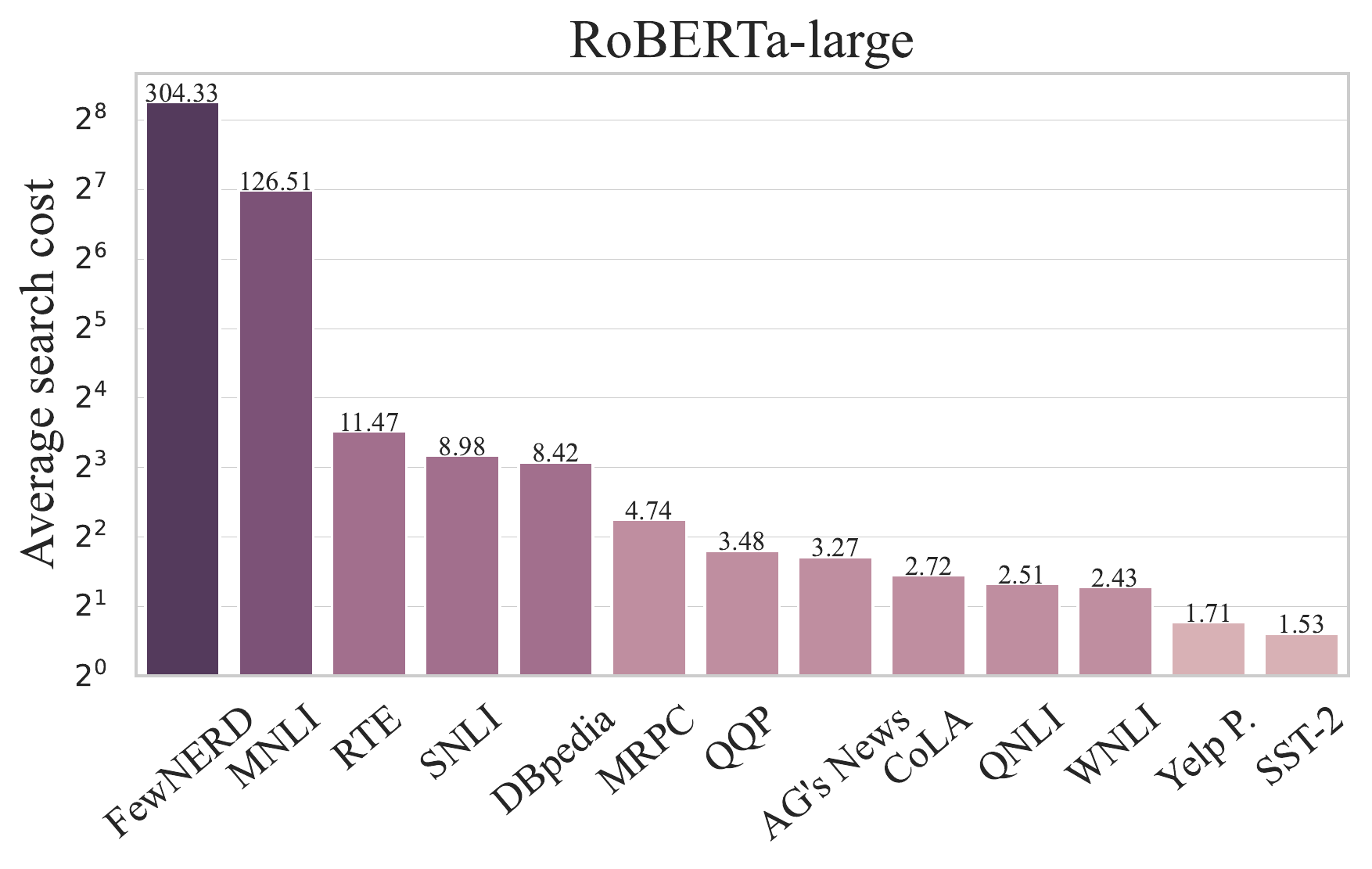}
% 		\caption{RoBETa-large}
    \end{minipage}
    % }
    % \subfigure[]{
    \begin{minipage}{0.98\linewidth}
		\centering
		\includegraphics[width=0.9\linewidth]{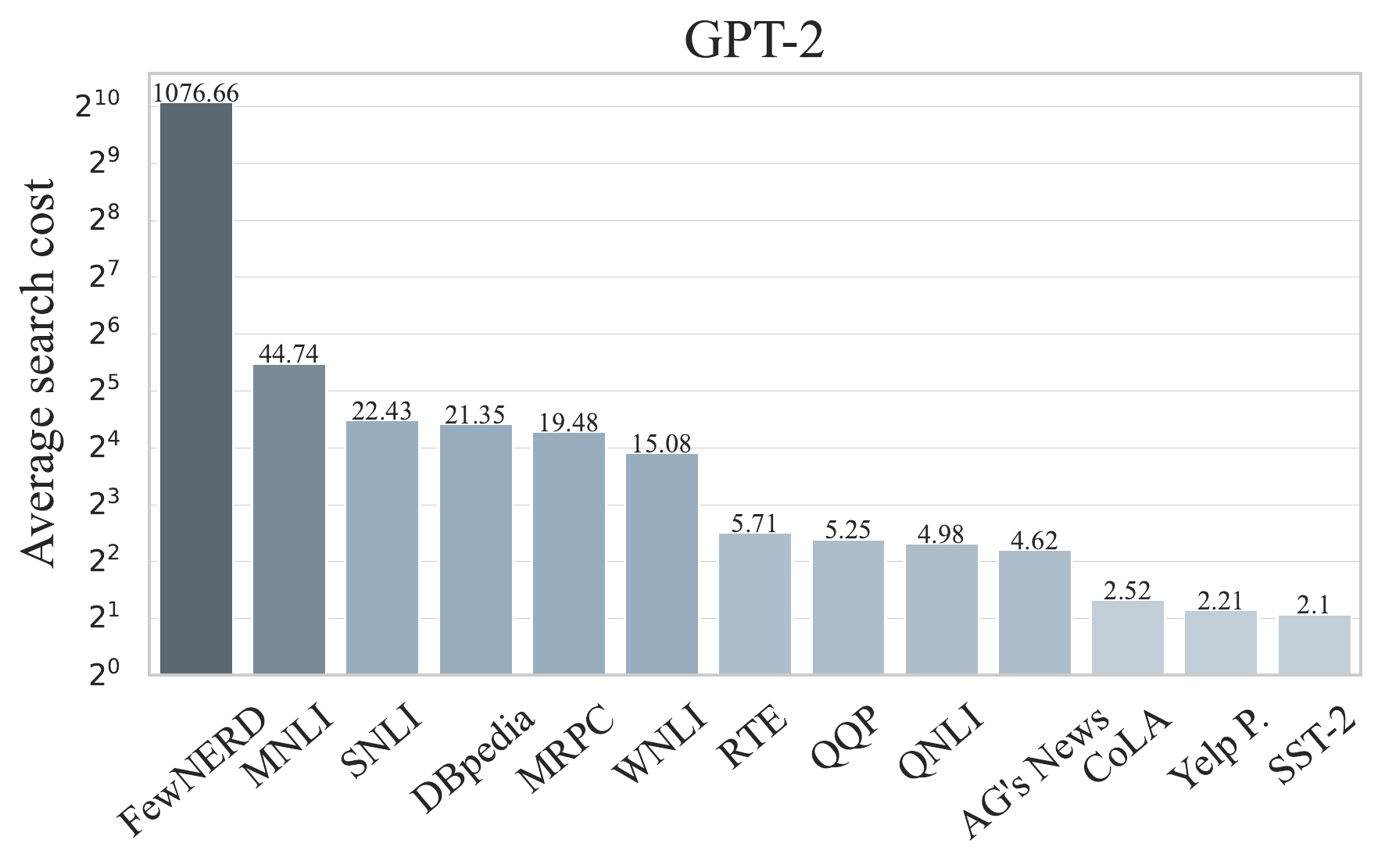}
% 		\caption{GPT-2}
    \end{minipage}
    % }
    \caption{The average search cost (number of API calls for each instance) on each dataset.}
    \label{fig:search}
    \vspace{-0.3cm}
% \vspace{-0.2cm}
\end{figure}

% \vspace{-0.1cm}

\subsection{Search Cost Analysis}
\label{sec:search_cost}
As aforementioned, the searching space in our experiment is $|\mathcal{T}|=76725$, however, the practical cost to find a lottery prompt for one data point is significantly lower than the budget. As shown in Figure~\ref{fig:search}, the average search cost for each instance does not exceed 30 API calls on most datasets for both PLMs. In this section, we show that search cost correlates with data difficulty and model capacity with further analysis.

\noindent \textbf{Task Difficulty.} As shown in Figure~\ref{fig:search}, searching for a lottery prompt for a multi-class classification problem is more costly. The 66-class typing dataset Few-NERD requires a significantly higher search budget than the rest of the datasets, most of which only contain 2 or 3 classes. 
Another reasonable observation is that single sentence classification tasks are generally easier than tasks involving sentence pairs. As mentioned in the next part, it may be attributed to the designing of prompt format and label words.
% However, judging the acceptibility of one sentence may still be hard, possibly because sentence acceptibility is implicitly encoded in PLMs during language modeling pretraining phase and is too elusive to be prompted. 
Meanwhile, NLI tasks with mixed domains are probably the most difficult sentence-pair tasks, given that MNLI, RTE, and SNLI are more costly than paraphrase tasks and other domain-specific NLI datasets.
% It indicates that when there are no obvious patterns, reasoning is more difficult for PLMs than direct semantics matching.
Comparing across models, the auto-regressive model (GPT-2) generally takes more searches than the auto-encoding model (RoBERTa-large). Despite the differences in individual datasets, they show similar trends, which can \textbf{roughly reflect how difficult the dataset is for PLMs}.

\noindent \textbf{Hard Instances.} Beyond task difficulty, we are also interested in some of the hard instances, i.e. instances that require a significant number of searches or fail to match any lottery prompt in the given search space.
We gather the 5 instances that require the most searches or ultimately observe a failure in searching from both PLMs. The examples from 3 datasets are presented in Appendix Table~\ref{tab:hard_example}.
It can be seen that for SST-2, the presented cases are intuitively difficult, as many of them involve vague expressions and complex reasoning that can be misleading.
On the other hand, the hard cases in MNLI and SNLI seem more counter-intuitive. Most ``entailment'' cases have considerable vocabulary overlap between the premise and hypothesis statements. The three failed cases are short sentences with almost identical expressions. We believe it is the negative effect from prompt template and label word chosen. For MNLI, both the high-lighted cases contain negation auxiliaries that rarely follow a ``Yes'' statement. This tendency drives the PLMs to always favor the choice of ``No'', which leads to erroneous prediction. The effect of negation has also been studied with standard PLM finetuning and proved to be a challenge~\citep{hossain-etal-2022-analysis, arian-etal-2021-understanding}.
The analysis shows that although for most instances, the lottery prompts can be easily found, \textbf{the prompting method is still disadvantaged when it comes to complex text that requires advanced understanding ability}. Also, prompting method is sensitive to verbalizer designs and \textbf{can be easily influenced by statistical correlation between label words and input texts.}

\noindent \textbf{Impact of Model Size and Pre-training.}
\label{sec:pretrain}
To explore the effect of model capacity on the easiness to search for lottery prompts, we conduct the same searching process as described in \cref{sec:existence} on AG's News and FewNERD with PLMs of different scales and pre-training status. Specifically, we use GPT-2, GPT-2-medium, GPT-2-large and GPT-2-xl for model size ablation and RoBERTa-base pre-trained for 5000$\sim$100000 steps for pre-training ablation, respectively. Figure~\ref{fig:pretrain} shows the variation of search success rate and average search cost per instance. 
For models of different scale, the success rate is similar but the search cost consistently decreases as models scale up, which shows that large PLMs generally have a larger feasible solution space for specific instances.
Meanwhile, finding lottery prompts for PLM at their early pre-training stage is much harder. As the pre-training progresses, a significant reduction in search cost and increase in success rate follow. 
This indicates that \textbf{the existence of lottery prompts is not merely a stroke of luck, but a consequence of pre-training that expands model capacity and can be further strengthened as PLMs scale up.}

\begin{figure}[htbp]
    \centering
    \includegraphics[width=0.48\textwidth]{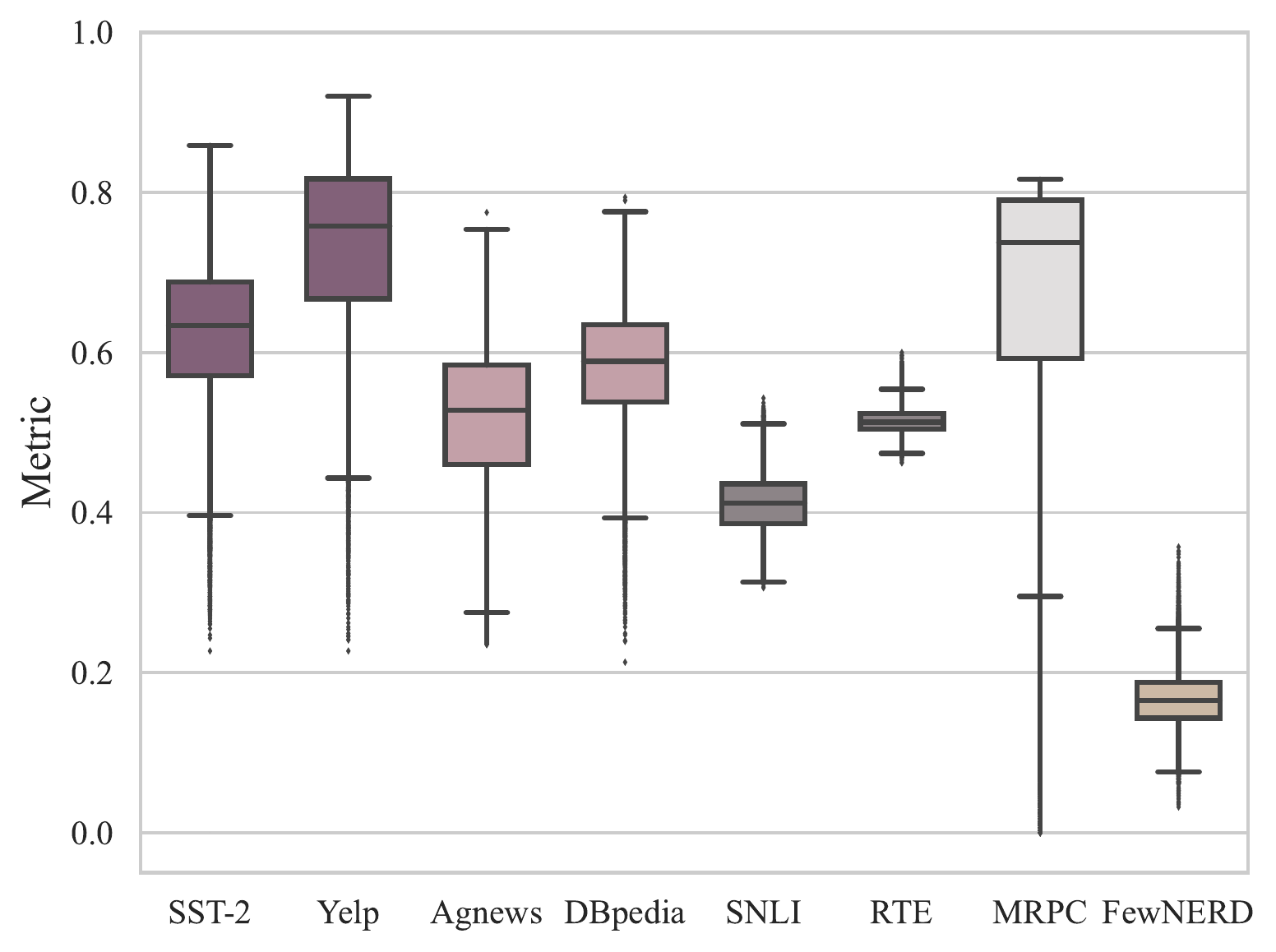}
    \caption{Prompt performance and variation on each dataset using RoBERTa-large. The vertical axis represents the metric of each prompt over $\mathcal{X}_\text{train}$. MRPC uses f1 metric, and others use accuracy.}
    \label{fig:template}
    % \vspace{-0.3cm}
\end{figure}

\begin{table*}[htbp!]
    \centering
    \scalebox{0.77}{
    \begin{tabular}{lll}
        \toprule
        \textbf{Dataset} & \textbf{Top-5 Prompts} & \textbf{Metrics} \\
        \midrule
        SST-2 & he work just, I find very, I find really, help are for, she work just & 85.9, 85.6, 85.2, 84.6, 84.0 \\
        % \midrule
        Yelp P. & look place really, you place also, look was also, I were very, they place also & 92.0, 91.3, 91.3, 91.2, 91.2 \\
        \midrule
        SNLI & I get really, I like through, I said always, keep love through, you found that & 56.9, 56.0, 55.8, 55.8, 55.7 \\
        RTE & keep like always, way think such, life think same, end think such, end like always & 60.0, 59.7, 59.6, 59.6, 59.4 \\
        \midrule
        MRPC & money had very, something had very, I been very, help had very, life had very & 70.9, 70.5, 70.4, 70.4, 70.2 \\
        \midrule
        AG's News & lot say on, I said other, time think other, state say on, you think other & 79.7, 78.8, 78.1, 78.0, 77.3 \\
        DBpedia & you said such, something know then, life make of, home said such, information is that & 87.5, 86.6, 86.0, 85.9, 85.8 \\
        \bottomrule
    \end{tabular}}
    \caption{An example of Top-5 prompts over 1000 training instances for each dataset and their individual performance on training sets. The model used is RoBERTa-large.}
    \label{tab:top_prompt_example}
\end{table*}

\vspace{-0.5cm}
% \subsection{Further Analysis}
\subsection{The Strong Prompts} 
\label{sec:strong_prompt}
After searching for lottery prompts for all instances, we are interested in if there are ``strong prompts'' among them, i.e., prompts that perform well on the whole $\mathcal{X}_\text{train}$. 
% could perform well on the whole $\mathcal{X}_\text{train}$. 
We measure the performance of each prompt over $\mathcal{X}_\text{train}$ with standard metrics on some representative datasets from each task category. The metric statistics and variation of all prompts are shown in Figure~\ref{fig:template}. 
\textbf{It could be concluded that for all datasets, there are a handful of ``strong prompts'' that can perform satisfactorily on the dataset.} Note that despite having altogether 66 classes, the best-performing prompt almost achieves an accuracy of 0.4 on Few-NERD. 
Meanwhile, different tasks show distinct patterns. 
Text classification tasks with single sentence are more sensitive to prompt choice and often observe larger performance variation over the prompt space. 
For SST-2, while the best-performing prompt reaches an accuracy of 0.8, the worst prompts can barely get to 0.3. For NLI tasks, prompt performance is more stable however mediocre.

To inspect into the linguistic characteristics of the strong prompts, we present the top-5 prompts for some of the representative datasets and their corresponding metrics on the training set of 1000 instances in Table~\ref{tab:top_prompt_example}. 
While many prompts may not seem syntactical on the whole, certain linguistic characteristics can still be identified, which fit with our language intuition, both syntactically and semantically, and reveal some of the most contributive words in prompts for distinctive datasets. For example, the top prompts for the sentiment analysis task are compatible with chosen label words. Adverbs that enhance the statement (e.g. just, really, very) appear frequently in sentiment analysis tasks. For topic classification, the words like ``other'' and ``such'' naturally lead to nominal label words like ``sports'' and ``artist''. 
As for natural language inference task, although language entailment is subjective, it is common that personal pronouns are often involved when we express our opinions on entailment, like "I think it means", "Do you think", etc. Therefore appearance pf pronouns is in top prompts is reasonable.
Meanwhile, we do observe that good prompts are not always interpretable. It may imply that the PLM's internal language ability and understanding deviates from human beings, which is why prompt engineering is important. 
Above all, we see that \textbf{``strong prompts'' do exist and they are equipped with distinct linguistic features depending on label words and task type.}
% The most important indication from the results is that some strong prompts could yield significant performance on $\mathcal{X}_\text{train}$, and such prompts could be obtained without any optimization process. 
% At this point, a natural question arises: 
% \textit{How to generalize these strong prompts to unseen (test) data?}

% Moreover, many of the top-performing prompts fit with our language intuition, both syntactically and semantically. For example, the top prompts for sentiment analysis task are compatible with chosen label words. Adverbs that enhance the statement (e.g. Just, really, very) appear frequently in sentiment analysis tasks. Also, different datasets observe distinct strong prompts patterns. For entity typing, the words like "other" and "such" naturally lead to noun-like words or category words. 

\begin{figure*}[!htbp]
    \centering
    \includegraphics[width=1\textwidth]{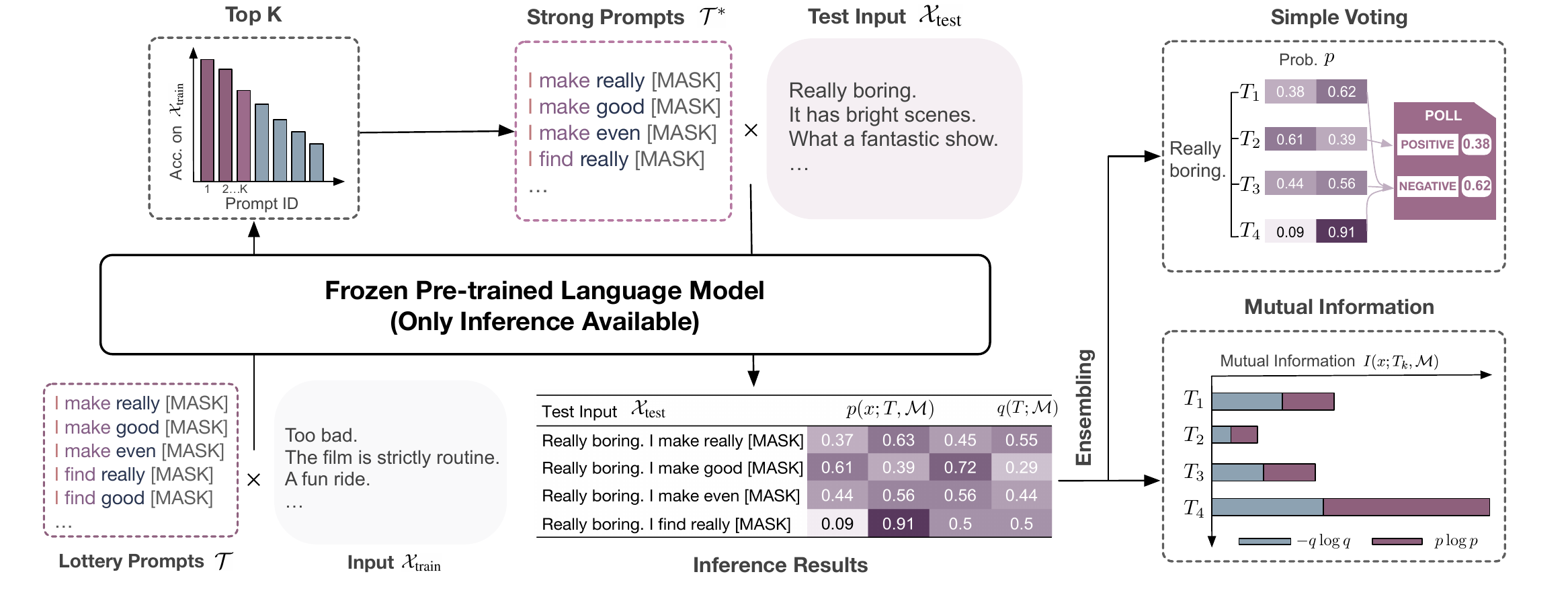}
    \caption{The complete process of searching for $\mathcal{T}^*$ and ensembling with $\Phi$.}
    \label{fig:stage2}
    % \vspace{-0.15cm}
\end{figure*}

%% file: generalization.tex
\section{Explore the Generalizability of Strong Prompts}
\label{sec:generaliza}

In ~\cref{sec:existence} and ~\cref{sec:emp_ana}, we have empirically verified that conditioned on a pre-trained model and a classification task, it is possible to find a lottery prompt for almost every data point, and that there are a handful of strong prompts that perform non-trivially on the training set.
In this section, we first describe the ensembling method and then present the generalization results.
% , targeting at research question Q4 and Q5. 
% Our considerations are:
% \vspace{-0.1cm}
% \begin{itemize*}
%     \item \looseness=-1 
%     % Since no optimization is conducted in the previous stage, there is no need to consider the problem of overfitting, that is, 
%     A strong prompt that performs well on the training set can naturally be assumed to perform relatively well on the test set. 
%     \item \looseness=-1 
%     % Although for the training set we can easily know which data points can be correctly classified with which prompts, this is unknown for the test set. 
%     It is empirically difficult to estimate which prompt will perform well on which instance for test data.
%     Therefore, we attempt to combine the predictions of selected strong prompts $\mathcal{T}^*=\{T_1, T_2, ..., T_t\} \in \mathcal{T}$. Under this circumstance, we investigate strategies to select $\mathcal{T}^*$ and ensemble the predictions of prompts in $\mathcal{T}^*$.
% \end{itemize*}
% We adopt different strategies to choose a set of templates for each dataset and use ensembling technique to predict on test set. 

\subsection{Prompt Ensembling Method}

\looseness=-1 We gather a set of feasible prompts $\mathcal{T}^*$ with the searching result on $\mathcal{X}_\text{train}$ and conduct inference with designed prompt ensembling method for each instance in $\mathcal{X}_\text{test}$.  
Since the choice of $\mathcal{T}^*$ is solely based on inference results on $\mathcal{X}_\text{train}$, the process uses no validation set.
Formally, given the selected prompts $\mathcal{T}^*=\{T_1, T_2, ..., T_t\} \subset \mathcal{T}$, the prediction for each data point $x \in \mathcal{X}_\text{test}$ is presented as
\begin{equation}
% \vspace{-0.1cm}
    p(x;\mathcal{T}^*, \mathcal{M}) = \Phi(p_1, p_2, ..., p_t),
\label{eq:vote}
% \vspace{-0.1cm}
\end{equation}
where $p_k = p(x;T_k, \mathcal{M})$ and is calculated as equation~\ref{eq:p}, and $\Phi$ is the ensembling method used.
With the assumption that strong prompts over $\mathcal{X}_\text{train}$ are also expected to perform well on $\mathcal{X}_\text{test}$, these best-performing prompts are regarded as the most reliable for predicting the unseen data. So we take the top-k best-performing prompts over the training set as $\mathcal{T}^*$. In the experiments, we empirically use $k=10$.
A naive ensembling method is to take the average output as the final prediction by simple voting, where $\Phi(p_1, p_2, ..., p_t) = \frac{1}{t} \sum_{k=1}^t p_k$.
% \begin{equation}
%     \Phi(p_1, p_2, ..., p_t) = \frac{1}{t} \sum_{k=1}^t p_k.
% \label{eq:simplevote}
% \end{equation}
While a more sophisticated strategy that echoes the spirit of ``lottery prompt'' is to select one most ``reliable'' prompt for each instance $x \in \mathcal{X}_\text{test}$. 
Intuitively, the more reliable a prompt $T$ is, the more confident the model $\mathcal{M}$ will be about instance $x$. Inspired by~\citet{sorensen2022information}, we measure the confidence with the mutual information between $x$ and $y$, $T$, which is defined by the reduction in entropy of predicted probability brought by $x$, 
% \vspace{-0.1cm}
\begin{equation}
\begin{aligned}
    I(x;T_k, \mathcal{M}) &= H(q|T_k(\cdot)) - H(p|T_k(x)) \\
    =\!\!- \sum_{i}&q_i(T_k;\mathcal{M})\log{q_i(T_k;\mathcal{M})}+\\
    \!\!\sum_{i}&p_i(x;T_k, \mathcal{M})\log{p_i(x;T_k, \mathcal{M})},
\end{aligned}
\label{eq:mi}
% \vspace{-0.1cm}
\end{equation}
where $q$ and $p$ are the predicted probability vectors as in Equation~\ref{eq:p}.
% It is expected that if a prompt is reliable for $x$, the uncertainty in prediction should be greatly reduced compared with making a prediction based on the prompt solely. 
So the overall objective is
\begin{equation}
\begin{aligned}
    T^{*} &= \arg\max_{T \in \mathcal{T}^*} I(x;T, \mathcal{M}), \\
    \Phi(p_1, &p_2, ..., p_t) = p(x;T^{*},\mathcal{M}). \\
\end{aligned}
\end{equation}
% where $p^* = p(x;T^{*(i)},\mathcal{M})$
Specifically, maximization of mutual information entails that a good prompt itself should contain no bias towards the label set, so $q$ should be close to a uniform distribution. On the other hand, a suitable prompt for a specific instance should induce an almost certain prediction on the desired class, corresponding to a near one-hot vector $p$.
Experiments show that under few-shot settings, our mutual-information-based ensembling strategy is more advantageous than direct simple voting (\cref{sec:exp}).
The complete searching and ensembling process is shown in Figure~\ref{fig:stage2}.

\begin{figure*}
    \centering
    \includegraphics[width=1.0\textwidth]{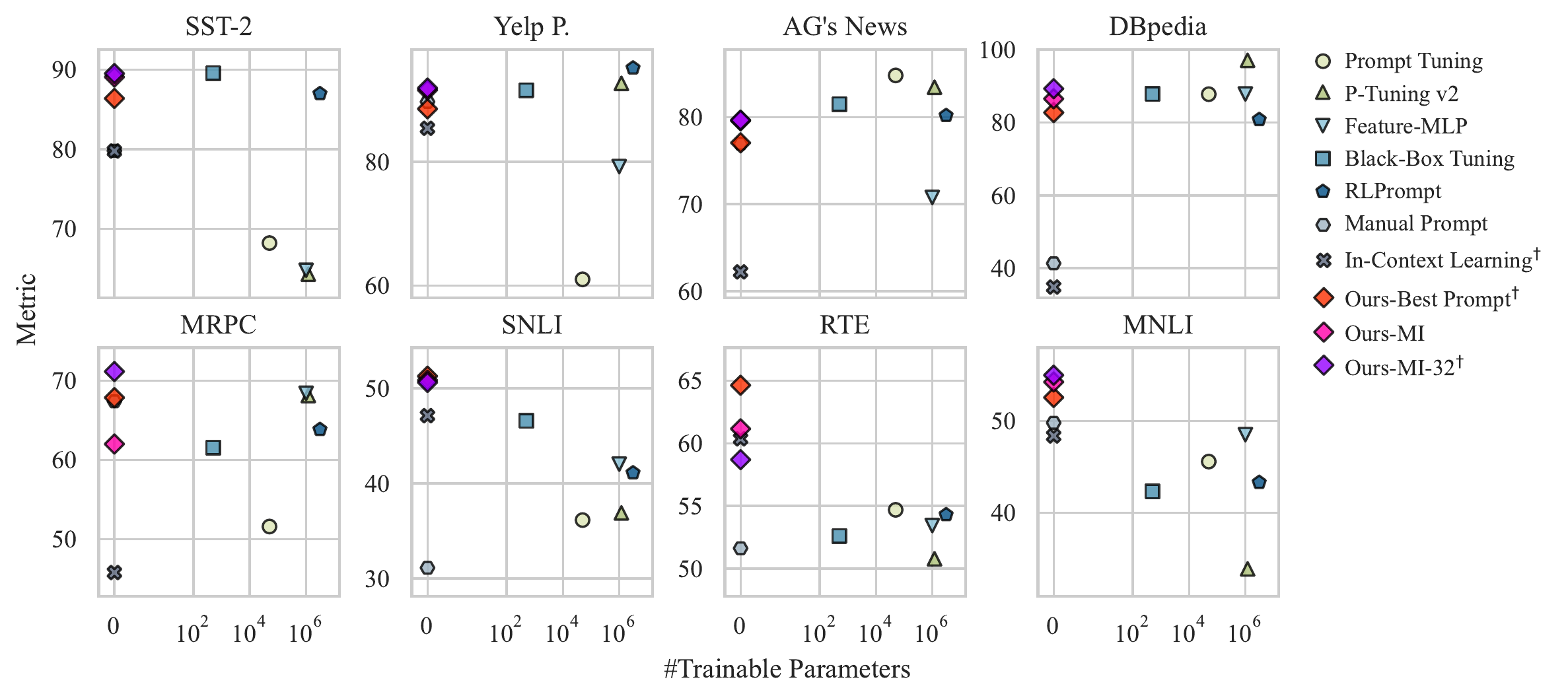}
    \caption{Performance on datasets under few-shot setting with RoBERTa-large. ``Best Prompt'' means directly using the top-1 performing searched prompt for test. $\dag$ means using 32-shot data as training set and no extra data as validation set, and manual prompt uses no training data. Other baseline methods use 16-shot data as training and validation set. Our method uses mutual-information-based prompt ensembling method (indicated as ``MI'') and average results over 5 runs are reported. The baseline results mainly follow~\citet{sun2022black}.}
    \label{fig:mainexp}
    % \vspace{-0.3cm}
\end{figure*} 

\subsection{In-Dataset Generalization}
\label{sec:exp}
\noindent \textbf{Experimental Setup.}
We comprehensively evaluate the generalizability of strong prompts on 8 representative datasets. 
Following previous works~\citep{sun2022black}, we conduct experiments under few-shot settings.
% We experiment under both 16-shot and 32-shot data as the training set as our method requires no validation set. The total seen labeled data number does not exceed 32-shot across all methods.
Specifically, we choose the top-10 prompts as $\mathcal{T}^*$ and obtain the final prediction and test metrics with mutual-information-based ensembling as $\Phi$ on the test set.
% we randomly select 16 or 32 instances for each class from the training set and obtain the predicted scores combined with each $T \in \mathcal{T}^*$. 
% Then we choose the top-10 prompts $\mathcal{T}^*$ and obtain the final prediction and evaluation metrics with specific $\Phi$ on the test set. 
% The original validation set is used as the test set following~\citep{sun2022black}.
% We choose comparable baselines that do not update the PLM parameters, including 1) Gradient-based methods: Prompt Tuning and P Tuning v2; 2) Optimization-based methods: Feature MLP, Black-box Tuning, and RLPrompt; and 3) Optimization-free methods: Manual Prompt, and In-Context learning. 
%%%%%%%%%%%%%%%%%%%%%%%%%%%%%%%%%%%%%%% add citation
% The results are mainly taken from \cite{sun2022black}. 
We keep the verbalizers aligned with \citet{sun2022black} for fair comparison.
% Since our method does not require any validation set, we provide results with both 16-shot training data and 32-shot training data, where validation data used in other optimization-based methods are added to our training data. 
The description of experimental details and baselines can be found in Appendix \ref{sec:expappendix}.

\noindent \textbf{Overall Results.}
Figure~\ref{fig:mainexp} shows the in-dataset generalization results on each dataset.
Overall, our method performs comparably to the existing baselines and requires the fewest trainable parameters. For some datasets, the searched strong prompts are shown to be more effective than baselines.
% Overall the proposed method performs the best among all methods.
It points to the fact that with a reasonable prompt search space and a few training instances, strong prompts can be identified and generalized effectively to unseen data. 
Best prompt on 32-shot data surprisingly overtakes many baselines. This, jointly with the mediocre performance of manual prompts, indicate that a human-comprehensible prompt may not always be the best choice for PLMs and may fail to probe a considerable amount of intrinsic knowledge in PLMs.
Meanwhile, the success of MI over best prompt shows that ensembling a set of strong prompts is beneficial. 
% plain textual prompt does work for PLMs and . 
Comparing across datasets, our method is more advantageous for harder tasks, including natural language inference (SNLI and RTE) and paraphrasing (MRPC). For single-sentence classification tasks, the improvement is minor. 
This finding fits with our intuition, as tasks involving two sentences often require more abstract abilities like reasoning and the contexts are more diverse across instances. Designing or optimizing for one unified prompt for such datasets is admittedly harder.
Above all, it is exciting that ensembling a proper set of prompts composed of textual tokens may surpass network optimization on a dataset in an era of pre-trained language models and points to the values of mining and tapping into an optimal usage of plain textual prompt. 

\begin{figure*}[!htbp]
    \centering
    \includegraphics[width=1.0\textwidth]{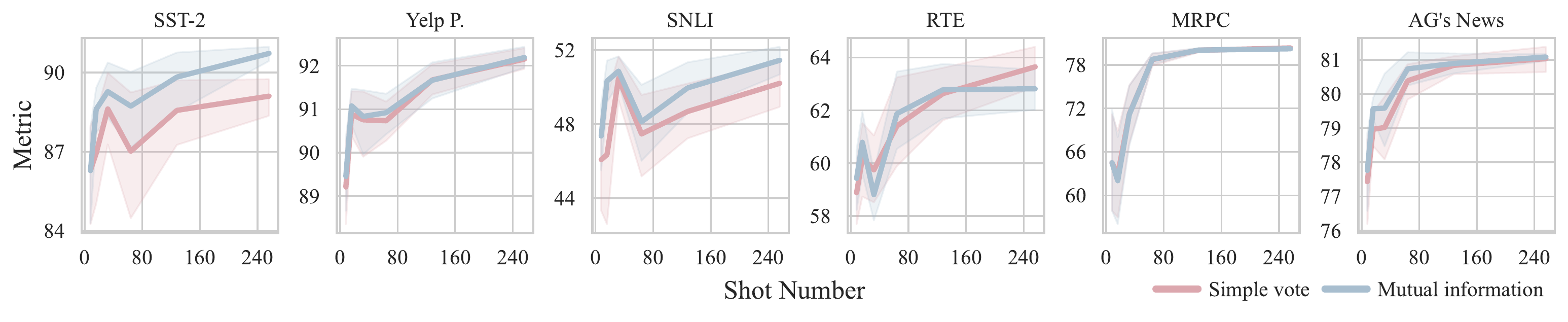}
    \caption{Average performance and standard deviation of the simple vote and mutual information under few-shot settings. We use 8, 16, 32, 64, 128, and 256-shot settings for training data. Top-10 prompts over each training set are adopted as $\mathcal{T}^*$. For each setting, the experiments are run with 5 different random seeds.}
    \label{fig:datasize}
    % \vspace{-0.3cm}
\end{figure*}

\noindent \textbf{Impact of Training Data Size.}
To further explore the property of our method, experiments are conducted under few-shot settings ranging from 8 shots to 256 shots with both simple voting and mutual-information-based ensembling.
As shown in Figure~\ref{fig:datasize}, we can see that performance varies a lot when different instances are sampled as the training set under low-shot settings. It suggests the importance of choosing the proper training data for our method. 
When more shots are provided, metrics get higher and variance gets smaller. As the data volume climbs up to 128 shots and 256 shots, the increase in metrics becomes minor for most datasets. 
It can also be concluded that for low-shot settings, mutual-information-based ensembling method yields higher results than simple voting. But as more training data are available, the gap is narrowed and the two ensembling strategies converge to similar levels.

\begin{table}[!htb]
    \centering
    \scalebox{0.85}{
    \begin{tabular}{llc}
        \toprule
        \textbf{Task} & \textbf{Setting} &  \textbf{Metrics} \\
        \midrule
        \multirowcell{2}{\textit{Sentiment} \\\textit{Analysis}} & SST-2 $\rightarrow$ Yelp P. & 90.27 ( $\color{red}{\text{1.58}\downarrow}$) \\
        & Yelp P. $\rightarrow$ SST-2 & 84.15 ( $\color{red}{\text{5.37}\downarrow}$) \\
        \midrule
        \multirowcell{4}{\textit{Natural}\\\textit{Language} \\\textit{Inference}} & RTE $\rightarrow$ SNLI & 40.48 ( $\color{red}{\text{10.13}\downarrow}$) \\
        & SNLI $\rightarrow$ RTE & 54.51 ($\color{red}{\text{4.19}\downarrow}$)\\
        \cmidrule{2-3}
        & MNLI $\rightarrow$ SNLI & 47.96 ($\color{red}{\text{2.65}\downarrow}$)\\
        & MNLI $\rightarrow$ RTE & 55.81 ($\color{red}{\text{2.89}\downarrow}$)\\
        \bottomrule
    \end{tabular}}
    \caption{Transferability test of $\mathcal{T}^*$ across datasets with similar tasks. Prompts are searched on 32-shot training data from source dataset and evaluated on test set of target dataset. Top-10 prompts are used as $\mathcal{T}^*$ and mutual-information-based strategy is used as $\Phi$.}
    % \vspace{-0.5cm}
    \label{tab:crossexp}
\end{table}

% \vspace{-0.2cm}
\subsection{Cross-Dataset Generalization}
\label{sec:cross}
We test the prompt transferbility across datasets with similar tasks under 32-shot setting.
Experiments are conducted on sentiment analysis and language inference tasks. We also use MNLI as the source dataset as many previous works do.
Table~\ref{tab:crossexp} shows that prompts chosen by our method can be transferable. While SST-2 and Yelp observe mutual transferability, transfering RTE to SNLI is relatively hard, which can be attributed to the inconsistency in class number. MNLI is shown to be a robust dataset for NLI task and the searched prompts perform satisfactorily on both RTE and SNLI. It is also in line with previous research findings that using prompts pretrained on MNLI could greatly boost performance on other NLI datasets.
Above all, the results demonstrate that our proposed strategy can effectively extract representative prompts for a specific kind of task, which can be further utilized to reduce search cost.
% shows that simpler tasks have better transferability among different datasets, with 5\% drop in SST2 and 2\% drop on Yelp. But complex tasks like NLI observe large performance drops up to 10\%.

%% file: appendix.tex
\section{Experimental Details}
\label{sec:expappendix}

\subsection{Experimental Settings}

\looseness=-1 We conduct experiments under few-shot settings on 8 datasets: SST-2, Yelp P., AG's News, DBpedia, MRPC, SNLI, RTE and MNLI. 
We experiment under both 16-shot and 32-shot data as the training set as our method requires no validation set. The total seen labeled data number does not exceed 32-shot across all methods.
% Specifically, we randomly select 16 or 32 instances for each class from the training set and obtain the predicted scores combined with each $T \in \mathcal{T}^*$. 
% Then we choose the top-10 prompts $\mathcal{T}^*$ and obtain the final prediction and evaluation metrics with specific $\Phi$ on the test set. 
The original validation set is used as the test set following~\citep{sun2022black}.
The detailed training and test set statistics for experiments in Figure~\ref{fig:mainexp} are shown in Table~\ref{tab:data}. 
All datasets are distributed under either CC BY license or CC BY-SA license, or subject to specified term of use. We have read and complied to the terms during experiments.
The label words used follow~\cite{sun2022black} and are the same across all methods.

\begin{table}[htb!]
    \centering
    \begin{tabular}{lcc}
        \toprule
        \textbf{Datasets} & \textbf{Classes} & \textbf{$|\mathcal{X}_\text{test}|$} \\
        \midrule
        SST-2 & 2 & 872 \\
        Yelp P. & 2 & 38000 \\
        AG's News & 4 & 7600 \\
        DBpedia & 14 & 70000 \\
        MRPC & 2 & 1725 \\
        SNLI & 3 & 10000 \\
        RTE & 2 & 277 \\
        MNLI & 3 & 9815 \\
        \bottomrule
    \end{tabular}
    \caption{Statistics of the  training and test set for experiments in Figure~\ref{fig:mainexp}.}
    \label{tab:data}
    \vspace{-0.5cm}
\end{table}

\subsection{Baselines}
\looseness=-1 
% Given that our method optimizes zero parameters, few existing methods are directly comparable. We choose some of the optimization-based methods that optimize only a small number of parameters instead.
We choose comparable baselines that do not update the PLM parameters, including 1) Gradient-based methods: Prompt Tuning and P Tuning v2; 2) Optimization-based methods: Feature MLP, Black-box Tuning, and RLPrompt; and 3) Optimization-free methods: Manual Prompt, and In-Context learning. The details are as follows:
\textbf{Prompt Tuning}~\citep{lester2021power} optimizes the continuous prompt at the input level.
\textbf{P-Tuning v2}~\citep{liu2021ptuning} is a variant of prompt tuning that pre-pends trainable parameters to each layer of the PLM and optimizes them in a multi-task setting.
\textbf{Feature-MLP} uses pre-trained features output by PLMs and train a lightweight classifier offline.
\textbf{Black-Box Tuning}~\citep{sun2022black} is a gradient-free method that optimizes the projected extra 500 parameters at the input layer with Covariance Matrix Adaptation Evolution Strategy.
\textbf{RLPrompt}~\citep{deng2022rlprompt} optimizes for discrete prompts with reinforcement learning.
\textbf{Manual Prompt} is a zero-shot method that directly uses a hand-crafted textual prompt for each dataset.
% \textbf{Best Prompt} uses the searched Top-1 prompt on training data directly.
\textbf{In-Context Learning}~\citep{brown2020language} is an optimization-free method that uses a few samples as demonstrations prepended to the test sample. 
Table~\ref{tab:method} lists the features and trainable parameter number of baselines and our method.
% %%%%%%%%%%%%% add table

\begin{table}[]
    \centering
    \scalebox{0.8}{
    \begin{tabular}{l|ccc}
        \toprule
        \textbf{Method} & \textbf{Gradients} & \textbf{Tuning} & \makecell{\textbf{\#Tunable}\\ \textbf{Param.}} \\
        \midrule
        Prompt Tuning & Yes & Yes & 50K  \\
        P-Tuning v2 & Yes &Yes & 1.2M  \\

        \midrule
        Feature-MLP & No & Yes & 1M  \\
        % Feature-BiLSTM & No & Yes & 17M  \\
        Black-Box Tuning & No & Yes & 500 \\
        RLPrompt & No & Yes & 3.1M  \\
        \midrule
        $\text{Manual Prompt}^\ddag$ & No & No & 0  \\
        $\text{In-Context Learning}^\dag$ & No & No & 0  \\
        $\text{Best Prompt}^\dag$ & No & No & 0 \\
        $\text{Ours}$ & No & No & 0 \\
        \bottomrule
    \end{tabular}}
    \caption{A summary of features of baselines methods and our method. ``Gradients'' refers to whether gradients of PLMs are required, and ``Tuning'' means whether updates of parameters are performed.}
    \label{tab:method}
\end{table}

\subsection{Implementation Details}
RoBERTa-large contains 354 million parameters and GPT-2 has 1.5 billion parameters. There is no extra parameter added in our method.
For each dataset, the experiments are run with 5 different random seeds, and the mean metrics are reported.
Most baseline results are taken from~\citet{sun2022black} and \citet{deng2022rlprompt}, while we re-run RLPrompt for MRPC and all baselines for MNLI with original code. 
% All methods in Figure use RoBERTa-large~\cite{liu2019roberta} as the PLM backbone. 
All experiments are conducted on NVIDIA A100 and GeForce RTX 3090 GPUs with CUDA.
% The complete search process for one dataset in \cref{sec:existence} takes about 10 hours with 4 GPUs to inference over every possible combination of $\{(x, y), T\} \in \mathcal{X}_\text{train} \times \mathcal{T}$. 
The search process in \cref{sec:exp} with 32-shot data takes about 2 hours with 40 GB maximum memory.
The test process takes 5$\sim$30 minutes depending on the size of $\mathcal{T}^*$ and $\mathcal{X}_\text{test}$. 
Our method is developed by OpenPrompt~\cite{ding-etal-2022-openprompt}, an open-source prompt-learning framework based on PyTorch~\cite{paszke2019pytorch}. The models are obtained from the Huggingface Transformers library~\cite{wolf-etal-2020-transformers}. 

% The word ``other'' naturally leads to discussions of objects of the same kind and is thus suitable for prompting the theme of input text.

\section{Efficiency Analysis}
\label{sec:efficiency}
The results reported in~\cref{sec:exp} all search through the whole prompt space $\mathcal{T}^*$, i.e. every combination of an instance and a prompt is covered. Since it would require up to 4 hours with a single NVIDIA A100, we seek to optimize the process by pruning the search space.
Our strategy is as follows: (1) randomly sample a batch of \textit{valid prompts} (in our experiments we use batch size 16) from $\mathcal{T}^*$ and apply them to the whole training set $\mathcal{X}_{\text{train}}$; (2) record the performance of each prompt word, i.e. if a prompt is \textit{``it was really''} and achieves 0.8 accuracy on $\mathcal{X}_{\text{train}}$, then for each word in the prompt (\textit{``it'', ``was'', ``really''}) 0.8 is recorded; (3) update the set of \textit{valid prompts}; (4) repeat until there is no remaining \textit{valid prompt}.
A \textit{valid prompt} means the average score of the three words is over a pre-defined threshold.
In our experiments, the threshold is set to 0.7 on SST-2 dataset and achieves a mean test accuracy of 87.6\%.

As shown in Table~\ref{tab:efficiency}, with the pruning strategy, the average time cost can be greatly reduced to 10 minutes with a still satisfying performance on test data. 
% The memory usage is especially large because our method allows for parallel computing where search on different prompts can be conducted simultaneously. 
In our experiments, prompts are randomly sampled and grouped into batches. We believe a better-designed and heuristically informed batching strategy will further boost the searching efficiency and test performance.

% \begin{table}[!htb]
%     \centering
%     \scalebox{0.9}{
%     \begin{tabular}{l|ccc}
%         \toprule
%         \textbf{Method} & \makecell{\textbf{Test} \\ \textbf{Acc.}} & \makecell{\textbf{Time}\\ \textbf{Cost}} & \makecell{\textbf{Memory} \\ \textbf{Cost}} \\
%         \midrule
%         Prompt Tuning & 72.6 & 15.9 mins & 5.3 GB \\
%         Feature-MLP & 63.8 & 7.0 mins & 2.8 GB \\
%         Feature-BiLSTM & 66.2 & 9.3 mins & 2.8 GB \\
%         Black-box Tuning & 89.4 & 10.1 mins & 3.0 GB \\
%         % Ours & & 240 mins & 19.1 GB \\
%         Ours+pruning & 87.6 & 10.3 mins & 19.1 GB \\
%         \bottomrule
%     \end{tabular}}
%     \caption{Training and searching time cost on SST-2. Following Table~\ref{tab:mainexp}, our method searches on 32-shot training data and other baselines are trained on 16-shot training data and evaluates on 16-shot validation data. The max sequence length is set uniformly to 47. Our method is run for 5 times on a single NVIDIA A100 and the mean time cost is reported.}
%     \label{tab:efficiency}
% \end{table}

\begin{table}[!htb]
    \centering
    \scalebox{1.0}{
    \begin{tabular}{l|ccc}
        \toprule
        \textbf{Method} &  \makecell{\textbf{Time} \textbf{Cost}}  \\
        \midrule
        Prompt Tuning  & 15.9 mins \\
        Feature-MLP  & 7.0 mins \\
        % Feature-BiLSTM  & 9.3 mins  \\
        Black-box Tuning & 10.1 mins \\
        % Ours & & 240 mins & 19.1 GB \\
        Ours & 10.3 mins \\
        \bottomrule
    \end{tabular}}
    \caption{Training and searching time cost on SST-2. Following Figure~\ref{fig:mainexp}, our method searches on 32-shot training data and other baselines are trained on 16-shot training data and evaluates on 16-shot validation data. The max sequence length is set uniformly to 47. Our method is run for 5 times on a single NVIDIA A100 and the mean time cost is reported.}
    \label{tab:efficiency}
\end{table}

\section{Details of Prompts and Label Words}
\label{sec:promptappendix}
Table~\ref{tab:search_format} displays the specific prompt format and label words used for searching for lottery prompt for each dataset with RoBERTa-large. For auto-regressive PLMs like GPT-2, the ``<mask>'' token are removed and the prediction of the next token by PLM will be extracted.

\section{Visualization of Words in Strong Prompts}
\begin{figure}[htb!]
    \centering
    \begin{minipage}{0.98\linewidth}
    \includegraphics[width=0.98\linewidth]{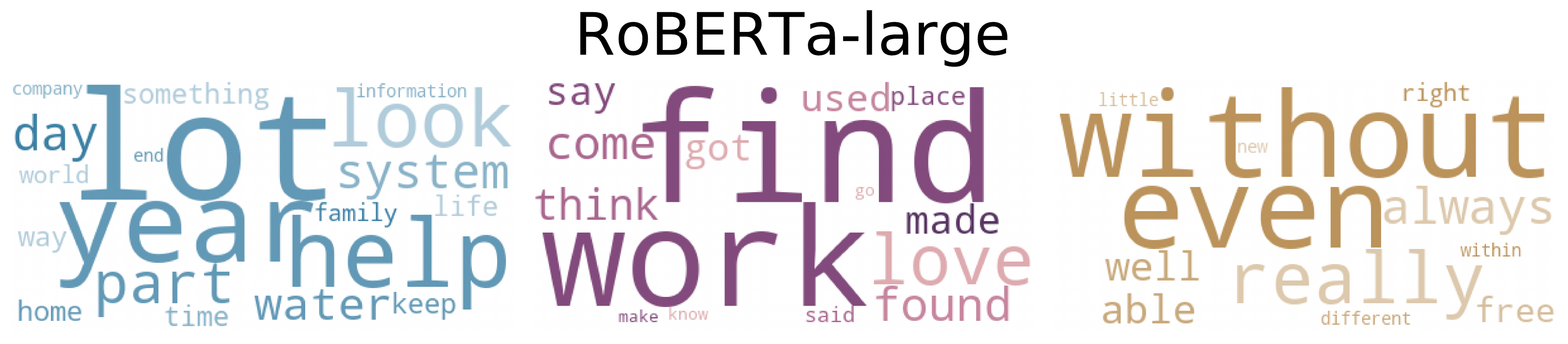}
    \end{minipage}
    \vspace{0.1cm}
    \begin{minipage}{0.98\linewidth}
    \includegraphics[width=0.98\linewidth]{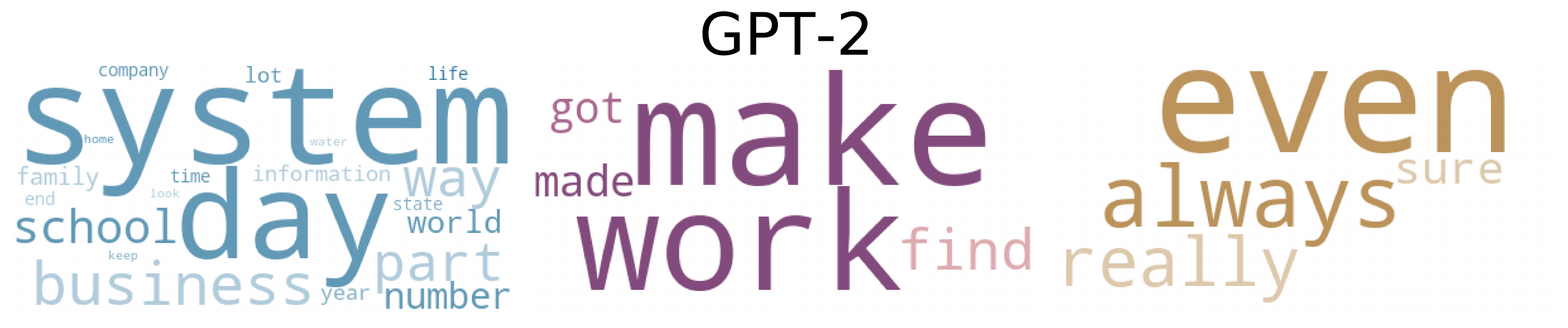}
    \end{minipage}
    \caption{\looseness=-1 Frequent words of 100 top-performing prompts at each position on SST-2. The 3 positions are [NOUN], [VERB], [PREP|ADJ|ADV] from left to right.}
    \label{fig:wordcloud}
\end{figure}

We also get the 100 best prompts out of $\mathcal{T}$ for SST-2, and visualize the frequent words at each position, as shown in Figure~\ref{fig:wordcloud}.
From the variation of words at each position, we can conclude that words more adjacent to the ``<mask>'' token has a larger impact on the prediction, which fits with our intuition.
In addition, GPT-2 demonstrates better fluency and interpretability compared to RoBERTa-large, as some high-frequency words found for RoBERTa-large like ``without'' are hard to comprehend.

\begin{table*}[htbp!]
    \centering
    \scalebox{0.85}{
    \begin{tabular}{lcl}
        \toprule
        \textbf{Dataset} & \textbf{Prompt} & \textbf{Label words} \\
        \midrule
        SST-2 & <Text> [Prompt] <mask> & great, bad \\
        Yelp P. & <Text> [Prompt] <mask> & great, bad \\
        \midrule
        CoLA & <Text> [Prompt] <mask> & reasonable, unreasonable \\
        \midrule
        SNLI & <Text1> [Prompt]? <mask>, <Text2> & Yes, Maybe, No \\
        RTE & <Text1> [Prompt]? <mask>, <Text2> & Yes, No \\
        MNLI & <Text1> [Prompt]? <mask>, <Text2> & Yes, Maybe, No\\
        QNLI & <Text1> [Prompt]? <mask>, <Text2> & Yes, No \\
        WNLI & <Text1> [Prompt]? <mask>, <Text2> & Yes, No \\
        \midrule
        MRPC & <Text1> [Prompt]? <mask>, <Text2> & Yes, No \\
        QQP & <Text1> [Prompt]? <mask>, <Text2> & Yes, No \\
        \midrule
        AG's News & <Text> [Prompt] <mask> & world, sports, business, technology \\
        DBpedia & <Text> [Prompt] <mask> & \makecell[l]{company, school, artist,  athlete, \\politics, transportation, building, river, \\village, animal, plant, album, \\film, book} \\
        \midrule
        Few-NERD & <Text> <Entity> [Prompt] <mask> & \makecell[l]{water, law, broadcast/program, media/newspaper, \\restaurant, artist/author, film, award, park, \\event, government/agency, person, educational/degree,\\ education, director, game, sports/facility, \\protest, car, language, airport, organization, \\building, location, athlete, show/organization, \\sports/league, geopolitical, scholar/scientist, library, \\hotel, road/railway/highway/transit, painting, hospital, \\election, written/art, religion, company, \\train, ship, attack/battle/war/military/conflict, sports/event, \\disaster, currency, weapon, living, sports/team, \\politician, god, political/party, music, \\art, actor, theater, biology, software, island, \\medical, disease, chemical, product, \\airplane, food, mountain, astronomy, soldier} \\
        \bottomrule
    \end{tabular}}
    \caption{The prompt format and label words used for each dataset. [Prompt] represents the sequence of ``[NOUN]  [VERB] [PREP|ADJ|ADV]''. For GPT-2, ``<Text1> [Prompt]? <mask>, <Text2> '' is changed into ``<Text1> <Text2> [Prompt]? <mask>''.}
    \label{tab:search_format}
\end{table*}

\begin{table*}[htb!]
    \centering
    \scalebox{0.76}{
    \begin{tabular}{lllc}
        \toprule
        \textbf{Datasets} & \multicolumn{2}{c}{\textbf{Instance Text}}& \textbf{Label} \\
        \midrule
        \multirow{8}{*}{SST-2} & \multicolumn{2}{l}{it falls far short of poetry , but} & negative \\
        \cmidrule{2-4}
        & \multicolumn{2}{l}{\makecell[l]{will be best appreciated by those willing to endure its extremely languorous rhythms , \\waiting for happiness}} & negative \\
        \cmidrule{2-4}
        & \multicolumn{2}{l}{expiration date} & negative \\
        \cmidrule{2-4}
        & \multicolumn{2}{l}{gut-wrenching , frightening war scenes since `` saving private ryan ''} & positive \\
        \cmidrule{2-4}
        & \multicolumn{2}{l}{sit through -- despite some first-rate performances} & positive \\
        \cmidrule{2-4}
        & \multicolumn{2}{l}{largely flat and uncreative} & negative \\
        \cmidrule{2-4}
        & \multicolumn{2}{l}{all of dean 's mannerisms and self-indulgence , but} & negative \\
        \cmidrule{2-4}
        & \multicolumn{2}{l}{\makecell[l]{if oscar had a category called best bad film you thought was going to be really awful but was n't}} & positive \\
        \midrule
        \multirow{8}{*}{MNLI} & \makecell[l]{It would be nice if more of the newcomers \\were artists, artisans, and producers, \\rather than lawyers and lobbyists, but head for head, \\I'll stack up Washington's intellectual capital \\against any competitor's.}& \makecell[l]{It would be nice if there were \\more lawyers instead of artistic people.} & contradiction \\
        \cmidrule{2-4}
        & \cellcolor{mypurple}{i just couldn't watch that much TV}& \cellcolor{mypurple}{I couldn't watch that much TV} & \cellcolor{mypurple}{entailment} \\
        \cmidrule{2-4}
        & \makecell[l]{yeah uh well we did well we did you know \\we really did i mean i just don't understand these \\people that think taking an RV and parking it \\and sitting inside and watching TV and having your \\microwave it's not camping}& \makecell[l]{I don't think it's camping \\if you hang out in an RV.} & entailment \\
        \cmidrule{2-4}
        & \makecell[l]{Of course}& \makecell[l]{Maybe.} & contradiction \\
        \cmidrule{2-4}
        & \cellcolor{myblue}{I think not!}& \cellcolor{myblue}{I do not think so.} & \cellcolor{myblue}{entailment} \\
        \cmidrule{2-4}
        & \makecell[l]{Exhibit 10 Adjustment Factors Used to \\Account for Projected Real Income Growth \\through 2010 and 2020}& \makecell[l]{See Exhibit 10 for Adjustment Factors Used to \\Account for Projected Real Income Growth \\through 2010 and 2020} & neutral \\
        \cmidrule{2-4}
        & \makecell[l]{In the dark of night, their aim must be true.}& \makecell[l]{Their aim must be accurate in the dark,\\ or else they will not succeed.} & neutral \\
        \cmidrule{2-4}
        & \makecell[l]{now we quit that about two years ago no three years ago \\when we got China mugs for everybody}& \makecell[l]{We stopped doing that three years ago, \\after we got everyone China mugs.} & entailment \\
        \midrule
        \multirow{10}{*}{SNLI} & \makecell[l]{Two men are playing a game of chess, \\one is standing and the other is sitting.}& \makecell[l]{A crowd watches a concert.} & contradiction \\
        \cmidrule{2-4}
        & \makecell[l]{A green jeep with men who are manning guns, \\with a crowd in the background on the street.}& \makecell[l]{Video game fans in cosplay outfits.} & contradiction \\
        \cmidrule{2-4}
        & \makecell[l]{A man has a pink ribbon around his arm.}& \makecell[l]{A guy with a strip of cloth around his bicep.} & entailment \\
        \cmidrule{2-4}
        & \makecell[l]{Large amounts of people walk around near \\a large, silver, reflective display.}& \makecell[l]{People are singing.} & contradiction \\
        \cmidrule{2-4}
        & \makecell[l]{Man playing the accordion on a sidewalk during the day.}& \makecell[l]{The Pope speed dials.} & contradiction \\
        \cmidrule{2-4}
        % & \makecell[l]{A green jeep with men who are manning guns, \\with a crowd in the background on the street.}& \makecell[l]{Video game fans in cosplay outfits.} & contradiction \\
        % \cmidrule{2-4}
        % & \makecell[l]{A man has a pink ribbon around his arm.}& \makecell[l]{A guy with a strip of cloth around his bicep.} & entailment \\
        % \cmidrule{2-4}
        
        & \makecell[l]{People walk and bike in front of a box office.}& \makecell[l]{People are carrying about their business \\nearby a box office} & entailment \\
        \cmidrule{2-4}
        & \makecell[l]{Three naked little boys are playing \\in a river and are covered in mud; \\one is standing up.}& \makecell[l]{the boys had no clothes on in the river} & entailment \\
        \cmidrule{2-4}
        & \makecell[l]{A person wearing a dark blue covered up \\attire from head to toe, with a mask and vest, \\holding a thin sword.}& \makecell[l]{Someone with a sword} & entailment \\
        \cmidrule{2-4}
        & \makecell[l]{Four children are in an industrial kitchen \\looking at a recipe with the ingredients \\on the table in front of them.}& \makecell[l]{Four people are in the kitchen} & entailment \\
        \cmidrule{2-4}
        & \cellcolor{mypurple}{Two guys getting a drink at a store counter.}& \cellcolor{mypurple}{two guys get a drink} & \cellcolor{mypurple}{entailment} \\
        % \cmidrule{2-4}
        \bottomrule
    \end{tabular}}
\caption{The most difficult instances for RoBERTa-large and GPT-2, measured by number of searches required to get the lottery prompt out of $\mathcal{T}$. Instances in \colorbox{mypurple} {purple} indicate failure to find a lottery prompt for GPT-2, and instances in \colorbox{myblue} {blue} are failure instances for RoBERTa-large.}
    \label{tab:hard_example}
\end{table*}

%% file: emnlp2021.bbl
\begin{thebibliography}{47}
\expandafter\ifx\csname natexlab\endcsname\relax\def\natexlab#1{#1}\fi

\bibitem[{Bowman et~al.(2015)Bowman, Angeli, Potts, and
  Manning}]{bowman-etal-2015-large}
Samuel~R. Bowman, Gabor Angeli, Christopher Potts, and Christopher~D. Manning.
  2015.
\newblock \href {https://doi.org/10.18653/v1/D15-1075} {A large annotated
  corpus for learning natural language inference}.
\newblock In \emph{Proceedings of EMNLP}, pages 632--642.

\bibitem[{Brown et~al.(2020)Brown, Mann, Ryder, Subbiah, Kaplan, Dhariwal,
  Neelakantan, Shyam, Sastry, Askell, Agarwal, Herbert-Voss, Krueger, Henighan,
  Child, Ramesh, Ziegler, Wu, Winter, Hesse, Chen, Sigler, Litwin, Gray, Chess,
  Clark, Berner, McCandlish, Radford, Sutskever, and
  Amodei}]{brown2020language}
Tom Brown, Benjamin Mann, Nick Ryder, Melanie Subbiah, Jared~D Kaplan, Prafulla
  Dhariwal, Arvind Neelakantan, Pranav Shyam, Girish Sastry, Amanda Askell,
  Sandhini Agarwal, Ariel Herbert-Voss, Gretchen Krueger, Tom Henighan, Rewon
  Child, Aditya Ramesh, Daniel Ziegler, Jeffrey Wu, Clemens Winter, Chris
  Hesse, Mark Chen, Eric Sigler, Mateusz Litwin, Scott Gray, Benjamin Chess,
  Jack Clark, Christopher Berner, Sam McCandlish, Alec Radford, Ilya Sutskever,
  and Dario Amodei. 2020.
\newblock \href
  {https://proceedings.neurips.cc/paper/2020/file/1457c0d6bfcb4967418bfb8ac142f64a-Paper.pdf}
  {Language models are few-shot learners}.
\newblock In \emph{Proceedings of NeurIPS}, volume~33, pages 1877--1901.

\bibitem[{Deng et~al.(2022)Deng, Wang, Hsieh, Wang, Guo, Shu, Song, Xing, and
  Hu}]{deng2022rlprompt}
Mingkai Deng, Jianyu Wang, Cheng-Ping Hsieh, Yihan Wang, Han Guo, Tianmin Shu,
  Meng Song, Eric~P. Xing, and Zhiting Hu. 2022.
\newblock \href {https://arxiv.org/abs/2205.12548} {Rlprompt: Optimizing
  discrete text prompts with reinforcement learning}.
\newblock \emph{arXiv preprint}, abs/2205.12548.

\bibitem[{Devlin et~al.(2019)Devlin, Chang, Lee, and
  Toutanova}]{devlin2018bert}
Jacob Devlin, Ming-Wei Chang, Kenton Lee, and Kristina Toutanova. 2019.
\newblock \href {https://doi.org/10.18653/v1/N19-1423} {{BERT}: Pre-training of
  deep bidirectional transformers for language understanding}.
\newblock In \emph{Proceedings of NAACL}, pages 4171--4186.

\bibitem[{Ding et~al.(2021{\natexlab{a}})Ding, Chen, Han, Xu, Xie, Zheng, Liu,
  Li, and Kim}]{ding2021prompt}
Ning Ding, Yulin Chen, Xu~Han, Guangwei Xu, Pengjun Xie, Hai-Tao Zheng, Zhiyuan
  Liu, Juanzi Li, and Hong-Gee Kim. 2021{\natexlab{a}}.
\newblock \href {https://arxiv.org/abs/2108.10604} {Prompt-learning for
  fine-grained entity typing}.
\newblock \emph{arXiv preprint}, 2108.10604.

\bibitem[{Ding et~al.(2022)Ding, Hu, Zhao, Chen, Liu, Zheng, and
  Sun}]{ding-etal-2022-openprompt}
Ning Ding, Shengding Hu, Weilin Zhao, Yulin Chen, Zhiyuan Liu, Haitao Zheng,
  and Maosong Sun. 2022.
\newblock \href {https://aclanthology.org/2022.acl-demo.10} {{O}pen{P}rompt: An
  open-source framework for prompt-learning}.
\newblock In \emph{Proceedings ACL}, pages 105--113.

\bibitem[{Ding et~al.(2021{\natexlab{b}})Ding, Xu, Chen, Wang, Han, Xie, Zheng,
  and Liu}]{ding-etal-2021-nerd}
Ning Ding, Guangwei Xu, Yulin Chen, Xiaobin Wang, Xu~Han, Pengjun Xie, Haitao
  Zheng, and Zhiyuan Liu. 2021{\natexlab{b}}.
\newblock \href {https://doi.org/10.18653/v1/2021.acl-long.248} {Few-{NERD}: A
  few-shot named entity recognition dataset}.
\newblock In \emph{Proceedings of ACL}, pages 3198--3213.

\bibitem[{Gao et~al.(2021)Gao, Fisch, and Chen}]{gao-etal-2021-making}
Tianyu Gao, Adam Fisch, and Danqi Chen. 2021.
\newblock \href {https://aclanthology.org/2021.acl-long.295} {Making
  pre-trained language models better few-shot learners}.
\newblock In \emph{Proceedings of ACL/IJCNLP}, pages 3816--3830.

\bibitem[{Gu et~al.(2022)Gu, Han, Liu, and Huang}]{gu-etal-2022-ppt}
Yuxian Gu, Xu~Han, Zhiyuan Liu, and Minlie Huang. 2022.
\newblock \href {https://aclanthology.org/2022.acl-long.576} {{PPT}:
  Pre-trained prompt tuning for few-shot learning}.
\newblock In \emph{Proceedings of ACL}, pages 8410--8423.

\bibitem[{Han et~al.(2021{\natexlab{a}})Han, Zhang, Ding, Gu, Liu, Huo, Qiu,
  Zhang, Han, Huang, Jin, Lan, Liu, Liu, Lu, Qiu, Song, Tang, Wen, Yuan, Zhao,
  and Zhu}]{HAN2021}
Xu~Han, Zhengyan Zhang, Ning Ding, Yuxian Gu, Xiao Liu, Yuqi Huo, Jiezhong Qiu,
  Liang Zhang, Wentao Han, Minlie Huang, Qin Jin, Yanyan Lan, Yang Liu, Zhiyuan
  Liu, Zhiwu Lu, Xipeng Qiu, Ruihua Song, Jie Tang, Ji-Rong Wen, Jinhui Yuan,
  Wayne~Xin Zhao, and Jun Zhu. 2021{\natexlab{a}}.
\newblock \href {https://doi.org/https://doi.org/10.1016/j.aiopen.2021.08.002}
  {Pre-trained models: Past, present and future}.
\newblock \emph{AI Open}.

\bibitem[{Han et~al.(2021{\natexlab{b}})Han, Zhao, Ding, Liu, and
  Sun}]{han2021ptr}
Xu~Han, Weilin Zhao, Ning Ding, Zhiyuan Liu, and Maosong Sun.
  2021{\natexlab{b}}.
\newblock \href {https://arxiv.org/abs/2105.11259} {Ptr: Prompt tuning with
  rules for text classification}.
\newblock \emph{arXiv preprint}, 2105.11259.

\bibitem[{Hossain et~al.(2022)Hossain, Chinnappa, and
  Blanco}]{hossain-etal-2022-analysis}
Md~Mosharaf Hossain, Dhivya Chinnappa, and Eduardo Blanco. 2022.
\newblock \href {https://aclanthology.org/2022.acl-short.81} {An analysis of
  negation in natural language understanding corpora}.
\newblock In \emph{Proceedings of ACL}, pages 716--723.

\bibitem[{Hosseini et~al.(2021)Hosseini, Reddy, Bahdanau, Hjelm, Sordoni, and
  Courville}]{arian-etal-2021-understanding}
Arian Hosseini, Siva Reddy, Dzmitry Bahdanau, R.~Devon Hjelm, Alessandro
  Sordoni, and Aaron~C. Courville. 2021.
\newblock \href {https://doi.org/10.18653/v1/2021.naacl-main.102}
  {Understanding by understanding not: Modeling negation in language models}.
\newblock In \emph{Proceedings of NAACL-HLT}, pages 1301--1312.

\bibitem[{Hu et~al.(2021)Hu, Ding, Wang, Liu, Li, and
  Sun}]{hu2021knowledgeable}
Shengding Hu, Ning Ding, Huadong Wang, Zhiyuan Liu, Juanzi Li, and Maosong Sun.
  2021.
\newblock \href {https://arxiv.org/abs/2108.02035} {Knowledgeable
  prompt-tuning: Incorporating knowledge into prompt verbalizer for text
  classification}.
\newblock \emph{arXiv preprint}, 2108.02035.

\bibitem[{Iyer et~al.(2017)Iyer, Dandekar, , and Csernai}]{iyer2017first}
Shankar Iyer, Nikhil Dandekar, , and Kornel Csernai. 2017.
\newblock First quora dataset release: Question pairs.

\bibitem[{Jiang et~al.(2020)Jiang, Xu, Araki, and
  Neubig}]{jiang-etal-2020-know}
Zhengbao Jiang, Frank~F. Xu, Jun Araki, and Graham Neubig. 2020.
\newblock \href {https://aclanthology.org/2020.tacl-1.28} {How can we know what
  language models know?}
\newblock \emph{TACL}, 8:423--438.

\bibitem[{Jin et~al.(2022)Jin, Lu, Zhang, and Zong}]{jin2022instance}
Feihu Jin, Jinliang Lu, Jiajun Zhang, and Chengqing Zong. 2022.
\newblock \href {https://arxiv.org/abs/2201.07126} {Instance-aware prompt
  learning for language understanding and generation}.
\newblock \emph{arXiv}, abs/2201.07126.

\bibitem[{Lester et~al.(2021)Lester, Al-Rfou, and Constant}]{lester2021power}
Brian Lester, Rami Al-Rfou, and Noah Constant. 2021.
\newblock \href {https://arxiv.org/abs/2104.08691} {The power of scale for
  parameter-efficient prompt tuning}.
\newblock \emph{arXiv preprint}, abs/2104.08691.

\bibitem[{Levesque(2011)}]{levesque2011logical}
Hector~J. Levesque. 2011.
\newblock \href {http://www.aaai.org/ocs/index.php/SSS/SSS11/paper/view/2502}
  {The winograd schema challenge}.
\newblock In \emph{Logical Formalizations of Commonsense Reasoning, Papers from
  the 2011 {AAAI} Spring Symposium}.

\bibitem[{Li and Liang(2021)}]{li2021prefix}
Xiang~Lisa Li and Percy Liang. 2021.
\newblock \href {https://aclanthology.org/2021.acl-long.353} {Prefix-tuning:
  Optimizing continuous prompts for generation}.
\newblock In \emph{Proceedings of ACL}, pages 4582--4597.

\bibitem[{Li et~al.(2022)Li, Che, Wang, Jiang, Xiong, and
  Chaturvedi}]{li-etal-2022-spe}
Yiyuan Li, Tong Che, Yezhen Wang, Zhengbao Jiang, Caiming Xiong, and Snigdha
  Chaturvedi. 2022.
\newblock \href {https://aclanthology.org/2022.emnlp-main.803} {{SPE:}
  symmetrical prompt enhancement for fact probing}.
\newblock In \emph{Proceedings of EMNLP}, pages 11689--11698.

\bibitem[{Liu et~al.(2021{\natexlab{a}})Liu, Ji, Fu, Du, Yang, and
  Tang}]{liu2021ptuning}
Xiao Liu, Kaixuan Ji, Yicheng Fu, Zhengxiao Du, Zhilin Yang, and Jie Tang.
  2021{\natexlab{a}}.
\newblock \href {https://arxiv.org/abs/2110.07602} {P-tuning v2: Prompt tuning
  can be comparable to fine-tuning universally across scales and tasks}.
\newblock \emph{arXiv preprint}, abs/2110.07602.

\bibitem[{Liu et~al.(2021{\natexlab{b}})Liu, Zheng, Du, Ding, Qian, Yang, and
  Tang}]{liu2021gpt}
Xiao Liu, Yanan Zheng, Zhengxiao Du, Ming Ding, Yujie Qian, Zhilin Yang, and
  Jie Tang. 2021{\natexlab{b}}.
\newblock \href {https://arxiv.org/abs/2103.10385} {Gpt understands, too}.
\newblock \emph{arXiv preprint arXiv:2103.10385}.

\bibitem[{Liu et~al.(2019)Liu, Ott, Goyal, Du, Joshi, Chen, Levy, Lewis,
  Zettlemoyer, and Stoyanov}]{liu2019roberta}
Yinhan Liu, Myle Ott, Naman Goyal, Jingfei Du, Mandar Joshi, Danqi Chen, Omer
  Levy, Mike Lewis, Luke Zettlemoyer, and Veselin Stoyanov. 2019.
\newblock \href {http://arxiv.org/abs/1907.11692} {Roberta: {A} robustly
  optimized {BERT} pretraining approach}.
\newblock \emph{arXiv preprint}, abs/1907.11692.

\bibitem[{Loper and Bird(2002)}]{loper2002nltk}
Edward Loper and Steven Bird. 2002.
\newblock \href {https://doi.org/10.3115/1118108.1118117} {Nltk: The natural
  language toolkit}.
\newblock In \emph{Proceedings of the ACL-02 Workshop on Effective Tools and
  Methodologies for Teaching Natural Language Processing and Computational
  Linguistics - Volume 1}, page 63–70.

\bibitem[{Paszke et~al.(2019)Paszke, Gross, Massa, Lerer, Bradbury, Chanan,
  Killeen, Lin, Gimelshein, Antiga et~al.}]{paszke2019pytorch}
Adam Paszke, Sam Gross, Francisco Massa, Adam Lerer, James Bradbury, Gregory
  Chanan, Trevor Killeen, Zeming Lin, Natalia Gimelshein, Luca Antiga, et~al.
  2019.
\newblock Pytorch: An imperative style, high-performance deep learning library.
\newblock \emph{Advances in neural information processing systems}, 32.

\bibitem[{Petroni et~al.(2019)Petroni, Rockt{\"a}schel, Riedel, Lewis, Bakhtin,
  Wu, and Miller}]{petroni-etal-2019-language}
Fabio Petroni, Tim Rockt{\"a}schel, Sebastian Riedel, Patrick Lewis, Anton
  Bakhtin, Yuxiang Wu, and Alexander Miller. 2019.
\newblock \href {https://aclanthology.org/D19-1250} {Language models as
  knowledge bases?}
\newblock In \emph{Proceedings of EMNLP-IJCNLP}, pages 2463--2473.

\bibitem[{Radford et~al.(2019)Radford, Wu, Child, Luan, Amodei, and
  Sutskever}]{radford2019language}
Alec Radford, Jeff Wu, Rewon Child, David Luan, Dario Amodei, and Ilya
  Sutskever. 2019.
\newblock \href
  {https://d4mucfpksywv.cloudfront.net/better-language-models/language_models_are_unsupervised_multitask_learners.pdf}
  {Language models are unsupervised multitask learners}.

\bibitem[{Rae et~al.(2021)Rae, Borgeaud, Cai, Millican, Hoffmann, Song,
  Aslanides, Henderson, Ring, Young et~al.}]{rae2021scaling}
Jack~W Rae, Sebastian Borgeaud, Trevor Cai, Katie Millican, Jordan Hoffmann,
  Francis Song, John Aslanides, Sarah Henderson, Roman Ring, Susannah Young,
  et~al. 2021.
\newblock \href {https://arxiv.org/abs/2112.11446} {Scaling language models:
  Methods, analysis \& insights from training gopher}.
\newblock \emph{arXiv preprint arXiv:2112.11446}.

\bibitem[{Sainz et~al.(2021)Sainz, de~Lacalle, Labaka, Barrena, and
  Agirre}]{sainz2021label}
Oscar Sainz, Oier~Lopez de~Lacalle, Gorka Labaka, Ander Barrena, and Eneko
  Agirre. 2021.
\newblock \href {https://arxiv.org/abs/2109.03659} {Label verbalization and
  entailment for effective zero- and few-shot relation extraction}.
\newblock \emph{arXiv preprint}, abs/2109.03659.

\bibitem[{Sanh et~al.(2021)Sanh, Webson, Raffel, Bach, Sutawika, Alyafeai,
  Chaffin, Stiegler, Scao, Raja, Dey, Bari, Xu, Thakker, Sharma, Szczechla,
  Kim, Chhablani, Nayak, Datta, Chang, Jiang, Wang, Manica, Shen, Yong, Pandey,
  Bawden, Wang, Neeraj, Rozen, Sharma, Santilli, F{\'{e}}vry, Fries, Teehan,
  Biderman, Gao, Bers, Wolf, and Rush}]{sanh2021multi}
Victor Sanh, Albert Webson, Colin Raffel, Stephen~H. Bach, Lintang Sutawika,
  Zaid Alyafeai, Antoine Chaffin, Arnaud Stiegler, Teven~Le Scao, Arun Raja,
  Manan Dey, M.~Saiful Bari, Canwen Xu, Urmish Thakker, Shanya Sharma, Eliza
  Szczechla, Taewoon Kim, Gunjan Chhablani, Nihal~V. Nayak, Debajyoti Datta,
  Jonathan Chang, Mike~Tian{-}Jian Jiang, Han Wang, Matteo Manica, Sheng Shen,
  Zheng~Xin Yong, Harshit Pandey, Rachel Bawden, Thomas Wang, Trishala Neeraj,
  Jos Rozen, Abheesht Sharma, Andrea Santilli, Thibault F{\'{e}}vry, Jason~Alan
  Fries, Ryan Teehan, Stella Biderman, Leo Gao, Tali Bers, Thomas Wolf, and
  Alexander~M. Rush. 2021.
\newblock \href {https://arxiv.org/abs/2110.08207} {Multitask prompted training
  enables zero-shot task generalization}.
\newblock \emph{arXiv preprint}, abs/2110.08207.

\bibitem[{Schick et~al.(2020)Schick, Schmid, and
  Sch{\"u}tze}]{schick-etal-2020-automatically}
Timo Schick, Helmut Schmid, and Hinrich Sch{\"u}tze. 2020.
\newblock \href {https://doi.org/10.18653/v1/2020.coling-main.488}
  {Automatically identifying words that can serve as labels for few-shot text
  classification}.
\newblock In \emph{Proceedings of COLING}, pages 5569--5578.

\bibitem[{Schick and Sch{\"u}tze(2021{\natexlab{a}})}]{schick2021exploiting}
Timo Schick and Hinrich Sch{\"u}tze. 2021{\natexlab{a}}.
\newblock \href {https://aclanthology.org/2021.eacl-main.20} {Exploiting
  cloze-questions for few-shot text classification and natural language
  inference}.
\newblock In \emph{Proceedings of EACL}, pages 255--269.

\bibitem[{Schick and Sch{\"u}tze(2021{\natexlab{b}})}]{schick2021just}
Timo Schick and Hinrich Sch{\"u}tze. 2021{\natexlab{b}}.
\newblock \href {https://aclanthology.org/2021.naacl-main.185} {It{'}s not just
  size that matters: Small language models are also few-shot learners}.
\newblock In \emph{Proceedings of NAACL}, pages 2339--2352.

\bibitem[{Shin et~al.(2020)Shin, Razeghi, Logan~IV, Wallace, and
  Singh}]{shin-etal-2020-autoprompt}
Taylor Shin, Yasaman Razeghi, Robert~L. Logan~IV, Eric Wallace, and Sameer
  Singh. 2020.
\newblock \href {https://aclanthology.org/2020.emnlp-main.346} {{A}uto{P}rompt:
  {E}liciting {K}nowledge from {L}anguage {M}odels with {A}utomatically
  {G}enerated {P}rompts}.
\newblock In \emph{Proceedings of EMNLP}, pages 4222--4235.

\bibitem[{Socher et~al.(2013)Socher, Perelygin, Wu, Chuang, Manning, Ng, and
  Potts}]{socher-etal-2013-recursive}
Richard Socher, Alex Perelygin, Jean Wu, Jason Chuang, Christopher~D. Manning,
  Andrew Ng, and Christopher Potts. 2013.
\newblock \href {https://aclanthology.org/D13-1170} {Recursive deep models for
  semantic compositionality over a sentiment treebank}.
\newblock In \emph{Proceedings of EMNLP}, pages 1631--1642.

\bibitem[{Sorensen et~al.(2022)Sorensen, Robinson, Rytting, Shaw, Rogers,
  Delorey, Khalil, Fulda, and Wingate}]{sorensen2022information}
Taylor Sorensen, Joshua Robinson, Christopher Rytting, Alexander Shaw, Kyle
  Rogers, Alexia Delorey, Mahmoud Khalil, Nancy Fulda, and David Wingate. 2022.
\newblock \href {https://aclanthology.org/2022.acl-long.60} {An
  information-theoretic approach to prompt engineering without ground truth
  labels}.
\newblock In \emph{Proceedings of ACL}, pages 819--862.

\bibitem[{Sun et~al.(2022)Sun, Shao, Qian, Huang, and Qiu}]{sun2022black}
Tianxiang Sun, Yunfan Shao, Hong Qian, Xuanjing Huang, and Xipeng Qiu. 2022.
\newblock \href {https://proceedings.mlr.press/v162/sun22e.html} {Black-box
  tuning for language-model-as-a-service}.
\newblock In \emph{Proceedings of ICML}, pages 20841--20855.

\bibitem[{Wang et~al.(2018)Wang, Singh, Michael, Hill, Levy, and
  Bowman}]{wang2018glue}
Alex Wang, Amanpreet Singh, Julian Michael, Felix Hill, Omer Levy, and Samuel
  Bowman. 2018.
\newblock \href {https://aclanthology.org/W18-5446} {{GLUE}: A multi-task
  benchmark and analysis platform for natural language understanding}.
\newblock In \emph{Proceedings of EMNLP Workshop {B}lackbox{NLP}: Analyzing and
  Interpreting Neural Networks for {NLP}}, pages 353--355.

\bibitem[{Warstadt et~al.(2019)Warstadt, Singh, and
  Bowman}]{warstadt2018neural}
Alex Warstadt, Amanpreet Singh, and Samuel~R. Bowman. 2019.
\newblock \href {https://doi.org/10.1162/tacl_a_00290} {Neural network
  acceptability judgments}.
\newblock \emph{TACL}, 7:625--641.

\bibitem[{Wei et~al.(2021)Wei, Bosma, Zhao, Guu, Yu, Lester, Du, Dai, and
  Le}]{wei2021fintuned}
Jason Wei, Maarten Bosma, Vincent~Y. Zhao, Kelvin Guu, Adams~Wei Yu, Brian
  Lester, Nan Du, Andrew~M. Dai, and Quoc~V. Le. 2021.
\newblock \href {https://arxiv.org/abs/2109.01652} {Finetuned language models
  are zero-shot learners}.
\newblock \emph{arXiv preprint}, abs/2109.01652.

\bibitem[{Williams et~al.(2018)Williams, Nangia, and
  Bowman}]{williams-etal-2018-broad}
Adina Williams, Nikita Nangia, and Samuel Bowman. 2018.
\newblock \href {https://doi.org/10.18653/v1/N18-1101} {A broad-coverage
  challenge corpus for sentence understanding through inference}.
\newblock In \emph{Proceedings of NAACL}, pages 1112--1122.

\bibitem[{Wolf et~al.(2020)Wolf, Debut, Sanh, Chaumond, Delangue, Moi, Cistac,
  Rault, Louf, Funtowicz, Davison, Shleifer, von Platen, Ma, Jernite, Plu, Xu,
  Le~Scao, Gugger, Drame, Lhoest, and Rush}]{wolf-etal-2020-transformers}
Thomas Wolf, Lysandre Debut, Victor Sanh, Julien Chaumond, Clement Delangue,
  Anthony Moi, Pierric Cistac, Tim Rault, Remi Louf, Morgan Funtowicz, Joe
  Davison, Sam Shleifer, Patrick von Platen, Clara Ma, Yacine Jernite, Julien
  Plu, Canwen Xu, Teven Le~Scao, Sylvain Gugger, Mariama Drame, Quentin Lhoest,
  and Alexander Rush. 2020.
\newblock \href {https://doi.org/10.18653/v1/2020.emnlp-demos.6} {Transformers:
  State-of-the-art natural language processing}.
\newblock In \emph{Proceedings of EMNLP: System Demonstrations}, pages 38--45.

\bibitem[{Wu et~al.(2022)Wu, Wang, Gu, Hou, Dong, Vydiswaran, and
  Ma}]{wu2022idpg}
Zhuofeng Wu, Sinong Wang, Jiatao Gu, Rui Hou, Yuxiao Dong, V.~G.~Vinod
  Vydiswaran, and Hao Ma. 2022.
\newblock \href {https://doi.org/10.48550/arXiv.2204.04497} {{IDPG:} an
  instance-dependent prompt generation method}.
\newblock \emph{arXiv preprint}, abs/2204.04497.

\bibitem[{Zhang et~al.(2021)Zhang, Li, Chen, Deng, Bi, Tan, Huang, and
  Chen}]{zhang2021differentiable}
Ningyu Zhang, Luoqiu Li, Xiang Chen, Shumin Deng, Zhen Bi, Chuanqi Tan, Fei
  Huang, and Huajun Chen. 2021.
\newblock \href {https://arxiv.org/abs/2108.13161} {Differentiable prompt makes
  pre-trained language models better few-shot learners}.
\newblock \emph{arXiv preprint}, abs/2108.13161.

\bibitem[{Zhang et~al.(2015)Zhang, Zhao, and LeCun}]{zhang2015character}
Xiang Zhang, Junbo Zhao, and Yann LeCun. 2015.
\newblock \href {https://arxiv.org/abs/1509.01626} {Character-level
  convolutional networks for text classification}.
\newblock In \emph{Proceedings of NeurIPS}, page 649–657.

\bibitem[{Zhao et~al.(2021)Zhao, Wallace, Feng, Klein, and
  Singh}]{zhao2021calibrate}
Zihao Zhao, Eric Wallace, Shi Feng, Dan Klein, and Sameer Singh. 2021.
\newblock \href {http://proceedings.mlr.press/v139/zhao21c/zhao21c.pdf}
  {Calibrate before use: Improving few-shot performance of language models}.
\newblock In \emph{Proceedings of ICML}, pages 12697--12706. PMLR.

\end{thebibliography}
